\documentclass[journal]{IEEEtran} 
\pdfoutput=1
\IEEEoverridecommandlockouts
\usepackage{graphicx}
                                                          
\usepackage{amsmath,amssymb,amsfonts}
\usepackage{tabularx}
\usepackage{graphicx}
\usepackage{adjustbox}
\usepackage{textcomp}
\usepackage{xcolor}
\usepackage{hyperref}
\usepackage{amsmath}
\usepackage[noend]{algpseudocode}
\usepackage{comment}
\usepackage[normalem]{ulem}
\usepackage[colorinlistoftodos]{todonotes}
\usepackage[framemethod=tikz]{mdframed}
\newtheorem{dfn}{Definition}
\newtheorem{asm}{Assumption}
\newtheorem{remark}{Remark}
\usepackage{soul}
\usepackage{stfloats}
\usepackage{wrapfig}

\usepackage{cuted}
\usepackage{capt-of}
 
\usepackage{psfrag}
\usepackage{pstool}
\usepackage{multirow}
\algnewcommand\algorithmicforeach{\textbf{for each}}
\algdef{S}[FOR]{ForEach}[1]{\algorithmicforeach\ #1\ \algorithmicdo}

\newcommand{\footrefc}[1]{%
    $^{\ref{#1}}$%
}

\DeclareUnicodeCharacter{0301}{\'{e}}

\usepackage{xspace}

\makeatletter

\DeclareRobustCommand\onedot{\futurelet\@let@token\@onedot}
\def\@onedot{\ifx\@let@token.\else.\null\fi\xspace}

\def\eg{\emph{e.g}\onedot} 
\def\ie{\emph{i.e}\onedot}

\usepackage{algorithm,algpseudocode,setspace}

\usepackage{etoolbox}
\makeatletter
\patchcmd{\@makecaption}
  {\scshape}
  {}
  {}
  {}
\makeatletter
\patchcmd{\@makecaption}
  {\\}
  {.\ }
  {}
  {}
\makeatother
\def\tablename{Table}

\font\bfmath=cmmib10
\mathchardef\Gamma="7100
\mathchardef\Delta="7101
\mathchardef\Theta="7102
\mathchardef\Lambda="7103
\mathchardef\Xi="7104
\mathchardef\Pi="7105
\mathchardef\Sigma="7106
\mathchardef\Upsilon="7107
\mathchardef\Phi="7108
\mathchardef\Psi="7109
\mathchardef\Omega="710A

\mathchardef\alpha="710B
\mathchardef\beta="710C
\mathchardef\gamma="710D
\mathchardef\delta="710E
\mathchardef\epsilon="710F
\mathchardef\zeta="7110
\mathchardef\eta="7111
\mathchardef\theta="7112
\mathchardef\iota="7113
\mathchardef\kappa="7114
\mathchardef\lambda="7115
\mathchardef\mu="7116
\mathchardef\nu="7117
\mathchardef\xi="7118
\mathchardef\pi="7119
\mathchardef\rho="711A
\mathchardef\sigma="711B
\mathchardef\tau="711C
\mathchardef\upsilon="711D
\mathchardef\phi="711E
\mathchardef\chi="711F
\mathchardef\psi="7120
\mathchardef\omega="7121
\mathchardef\epsilon="7122

\mathchardef\varepsilon="7122
\mathchardef\vartheta="7123
\mathchardef\varpi="7124
\mathchardef\varrho="7125
\mathchardef\varsigma="7126
\mathchardef\varphi="7127
\mathchardef\imath="717B
\mathchardef\jmath="717C

\def\smallbfW{{\raise1.5pt\hbox{\mbox{\boldmath $_W$}}}}

\def\red{\color{red}}

\def\my4psfrag#1#2#3#4#5#6#7#8{
        \begin{figure}[htp]
        \begin{center}
            \begin{tabular}[h]{c c}
              {\leavevmode{\includegraphics[width=#1truecm]{#2.eps}}}
              &
              {\leavevmode{\includegraphics[width=#1truecm]{#3.eps}}} \\
              {\leavevmode{\includegraphics[width=#1truecm]{#4.eps}}}
              &
              {\leavevmode{\includegraphics[width=#1truecm]{#5.eps}}}
         \end{tabular}
           \vspace{#6}
           \caption{#7}
           \label{#8}
        \end{center}
        \end{figure}
}

\def\mydouble4psfrag#1#2#3#4#5#6#7#8{
        \begin{figure*}[htp]
        \begin{center}
            \begin{tabular}[h]{c c}
              {\leavevmode{\includegraphics[width=#1truecm]{#2.eps}}}
              &
              {\leavevmode{\includegraphics[width=#1truecm]{#3.eps}}} \\
              {\leavevmode{\includegraphics[width=#1truecm]{#4.eps}}}
              &
              {\leavevmode{\includegraphics[width=#1truecm]{#5.eps}}}
         \end{tabular}
           \vspace{#6}
           \caption{#7}
           \label{#8}
        \end{center}
        \end{figure*}
}

\makeatother
\usepackage[belowskip=-7pt,aboveskip=10pt,font={small}]{caption}
\usepackage[font={small}]{subcaption}

\DeclareMathOperator{\LSR}{LSR}
\newcommand{\lsr}[1]{\LSR\text{-}L_{#1}}
\DeclareMathOperator{\VAE}{VAE}
\newcommand{\vae}[3]{\VAE_{#1}\text{-}#2\text{-}{#3}}

\DeclareMathOperator{\APN}{APN}
\newcommand{\apn}[3]{\APN_{#1}\text{-}#2\text{-}{#3}}

\definecolor{orange}{RGB}{179, 98, 0}
  
\definecolor{green}{RGB}{78, 196, 164}
  
\definecolor{col}{RGB}{100, 10, 164}

\makeatletter

\begin{document}

\title{Enabling Visual Action Planning for Object Manipulation through Latent Space Roadmap
 \thanks{This work was supported by the Swedish Research Council, Knut and Alice Wallenberg Foundationm, by the European Research Council (ERC-884807), by the European Commission (Project CANOPIES-101016906), and  by Dipartimento di Eccellenza granted to DIEI
Department, University of Cassino and Southern Lazio.}
}

\author{ Martina Lippi*$^{1,2}$, Petra Poklukar*$^{1}$, Michael C. Welle*$^{1}$, Anastasia Varava$^{1}$, \\Hang Yin$^{1}$, Alessandro Marino$^{3}$,   and Danica Kragic$^{1}$% <-this % stops a space
\thanks{*These authors contributed equally (listed in alphabetical order)}% <-this % stops a space
\thanks{ ${}^1$KTH Royal Institute of Technology, Stockholm, Sweden}%
\thanks{ ${}^2$Roma Tre University, Rome, Italy}%
\thanks{ ${}^3$University of Cassino and Southern Lazio, Cassino, Italy}%
}

\maketitle

 \begin{abstract}

 We present a framework for visual action planning of complex manipulation tasks with high-dimensional state spaces, focusing on manipulation of deformable objects. We propose a Latent Space Roadmap (LSR) for task planning which is a graph-based structure globally capturing the system dynamics in a low-dimensional latent space.  
 Our framework consists of three parts: (1) a Mapping Module (MM) that maps observations given in the form of images into a structured latent space extracting the respective states as well as generates observations from the latent states, (2) the LSR which  builds and connects clusters containing similar states in order to 
 find the latent plans between start and goal states extracted by MM, and (3) the Action Proposal Module that complements the latent plan found by the LSR with the corresponding actions. We present a thorough investigation of our framework on  simulated box stacking and rope/box manipulation  tasks, and a folding task executed on a real robot. 
 
\end{abstract}

\section{Introduction}\label{sec:intro}

In task and motion planning, it is common to assume that the geometry of the scene is given as input to the planner. In contrast, modern representation learning methods are able to automatically extract state representations directly from high-dimensional raw observations, such as images or video sequences~\cite{oh2015action}. 
This is especially useful in complex scenarios where explicit analytical modeling of states is challenging,  such as in manipulation of \emph{highly deformable} objects which is recently gaining increasing attention by the research community \cite{bench_RAL2020,Yin_scirob_eabd8803}.  In these manipulation tasks, the state of the object cannot be easily established in a unique manner as opposed to manipulation of rigid objects, where their configuration can be made analytically explicit.   
\newline

\begin{figure}[!th]
\begin{center}
\includegraphics[width=\linewidth]{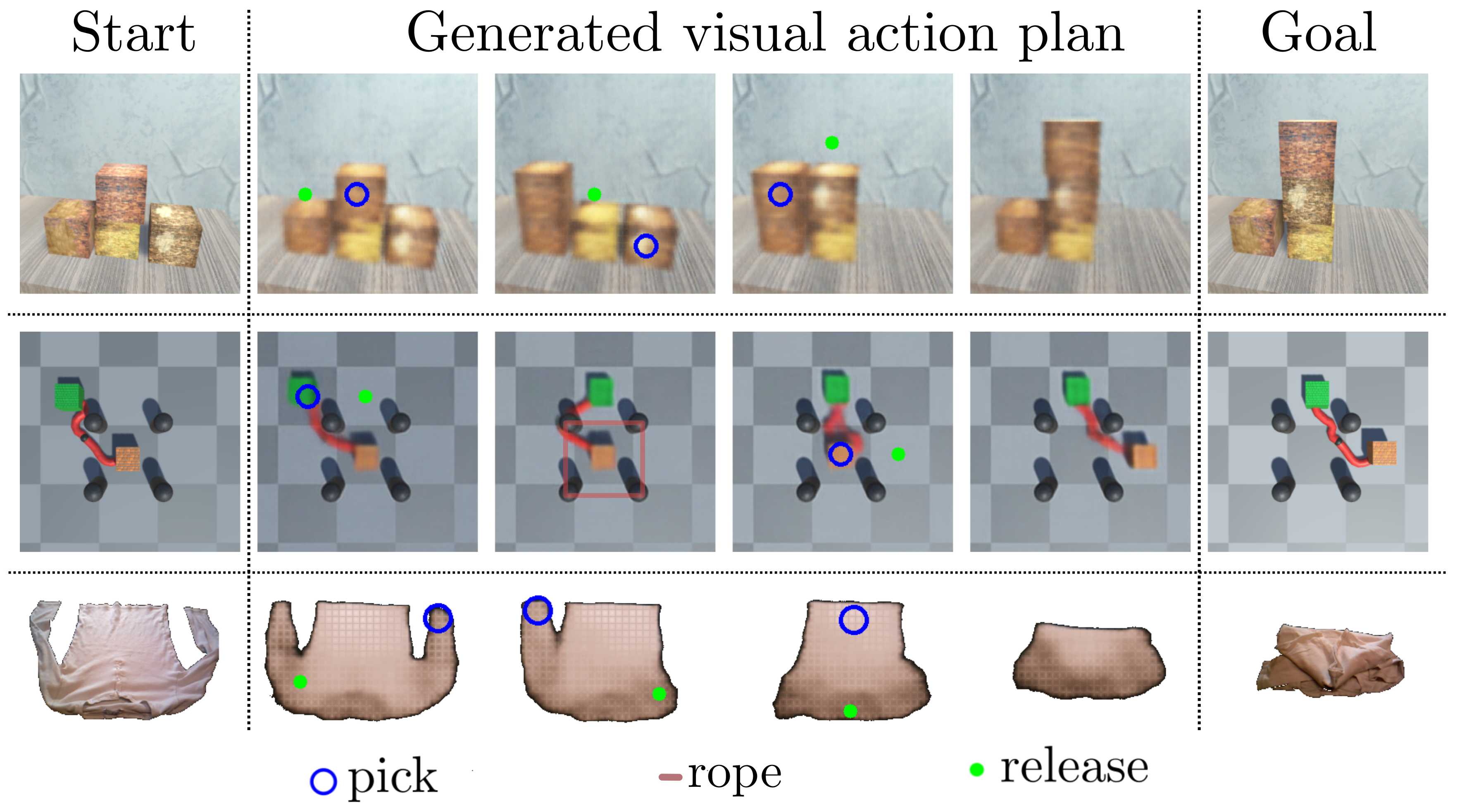}
\end{center}
\caption{Examples of visual action plans for a stacking task (top),  a rope/box manipulation task (middle) and a shirt folding task (bottom). }
\label{fig::vaps_fig1}
\end{figure}

\noindent{\bf Unsupervised State Representation Learning.}
Given raw observations, state representation learning is commonly performed in an unsupervised way using for example 
Autoencoders (AEs)~\cite{ballard1987modular} or Variational Autoencoders (VAEs)~\cite{kingma2013auto}.  In these frameworks, two neural networks -- an encoder and a decoder -- are jointly trained to embed the input observation into a low-dimensional latent space, and to reconstruct it given a latent sample.
The resulting latent space can be used as a low-dimensional representation of the state space, where the encoder acts as a map from a high-dimensional observation (an image) into the lower-dimensional state (a latent vector). 

However, to be useful for planning, it is desirable to have a particular structure in the latent space: states that are similar should be encoded close to each other, while different states should be separated. Such information does not always coincide with the similarity of the respective images: two observations can be significantly different with respect to a pixel-wise metric due to task-irrelevant factors of variation such as changes in the lighting conditions and texture, while the underlying state of the system (e.g., the shape and the pose of the objects) may be identical. The opposite is also possible: two observations may be relatively close in the image space, because the respective change in the configuration of the scene does not significantly affect the pixel-wise metric, while from the task planning perspective the two states are fundamentally different. \newline

\noindent{\bf Challenges of State Representation Learning for Planning.}
For planning, the system dynamics should also be captured in the latent space. We therefore identify three main challenges when modeling the state space representation for planning: \emph{i)} it needs to be low dimensional, while containing the relevant information from high-dimensional observations; \emph{ii)} it needs to properly reflect similarities between states; and \emph{iii)} it needs to efficiently capture feasible transitions between states allowing complex tasks such as deformable object manipulation.

In this work, we address \emph{i)} by extracting the low-dimensional states directly from images of the scene through a Mapping Module (MM). For this, we deploy a VAE framework and compare it to AE. 
We address \emph{ii)} by explicitly encouraging the encoder network to map the observations that correspond to different states further away from each other despite their visual similarity. This is done by providing a weak supervision signal: we record a small number of actions between observation pairs, and mark the observations as ``same" or ``different" depending on whether or not an action is needed to bring the system from one state to the successor one.
We use this action information in an additional loss term to structure the latent space accordingly.
Finally, we tackle  \emph{iii)} by building the Latent Space Roadmap (LSR), which is a graph-based structure in the latent space used to plan a manipulation sequence given a start and goal image of the scene. The nodes of this graph are associated with the system states, and the edges model the actions connecting them. For example, as shown in Fig.~\ref{fig::vaps_fig1}, these actions can correspond
to moving a box or a rope, or folding a shirt. We identify the regions containing the same underlying states using hierarchical clustering~\cite{mullner2011modern} which accounts for differences in shapes and densities of these regions. The extracted clusters are then connected using the weak supervision signals. Finally, the action specifics are obtained from the Action Proposal Module (APM). In this way, we capture the global dynamics of the state space in a data-efficient manner without explicit state labels, which allows us to learn a state space representation for complex \textit{long-horizon} tasks. \newline

\noindent{\bf Contributions.}
 Our contributions can be summarized as: 
 \begin{enumerate}
     \item We define the Latent Space Roadmap that  enables to generate visual action plans based on weak supervision; 
     \item  We introduce an augmented loss function with dynamic parameter  to favourably structure the latent space;
     \item We validate our framework on simulated box stacking tasks involving rigid objects, a combined rope and box manipulation task involving both deformable and rigid objects, and on a real-world T-shirt folding task involving deformable objects. Complete details can be found on the website\footnote{\label{fn:website}\url{https://visual-action-planning.github.io/lsr-v2/}}.
 \end{enumerate}
 
This work is an extensively revised  version  of our earlier conference paper \cite{ouriros}, where we first introduced the notion of Latent Space Roadmap. The main novelties of the present work with respect to \cite{ouriros} are: 
    \emph{i)} extension of the LSR building algorithm with an outer optimisation loop improving its performance,
    \emph{ii)} new training approach for the MM with a dynamic adjustment of the key hyperparameter used in the additional loss term, 
    \emph{iii)} large scale simulation campaigns investigating the effect of the additional loss term and hyperparameter choices,  
    \emph{iv)} restructuring of the framework into three main components leading to a more modular setup,
    \emph{v)} introduction of a more challenging box stacking task  and a task involving manipulation of a rope and two boxes, enabling a thorough ablation study on all components of our framework,
     \emph{vi)}  comparison with the state-of-the-art solutions in~\cite{savinov2018semiparametric} and~\cite{hafner2018learning} on the simulation tasks as well as comparison of the improved framework with its predecessor~\cite{ouriros} on the T-shirt folding task performed on a real robot,
    \emph{vii)} comparison with other potentially suitable clustering algorithms used to build the LSR,
    \emph{viii)} comparison of VAE and AE for the mapping module, 
    \emph{ix)} comparison of different realizations of the APM.

\section{Related Work}\label{sec:related-work}

Methods for planning in complex scenarios in which
the system state cannot be analytically established can be divided into two main categories based on where the planning is performed: 
\emph{i)} directly in a high-dimensional image space and \emph{ii)} in low-dimensional latent space. 
Belonging to \emph{i)}, a visual foresight framework was designed
in~\cite{finn2017deep} where a video prediction model based on Long-Short Term Memory blocks 
was employed to predict stochastic pixel flow from frame to frame. Trained on video, action and state sequences,  the model provides
an RGB prediction of the scene that  is
then used to perform visual model predictive control. The data was collected using ten identical real world setups with different camera angles.  To tackle long-horizon tasks, Reinforcement Learning (RL) combined with graph search over replay buffer was proposed in~\cite{eysenbach2019search} and validated with a visual navigation task.
 Planning in the image space has also been successfully applied to deformable objects as in~\cite{wang2019learning}, where   the manipulation of a rope from an initial start state to a desired goal state was analyzed. In particular,
a visual foresight plan is produced containing the intermediate steps to deform the rope using a Context Conditional Causal InfoGAN ($C^3$IGAN).  To this aim, the results of~\cite{nair2017combining} were exploited where $500$ hours worth of data collection  were used to learn the rope inverse dynamics.

 To mitigate the time burden of collecting data on real robots, simulators with deformable objects have also been employed, for example, in \cite{fabric_vsf_2020}, where 
a custom fabric simulator \cite{seita2020deep} was used to learn fabric dynamics building on the visual foresight model \cite{finn2017deep}. The learned dynamic models  are reusable and can be applied to different tasks given a single image goal-conditioned policy.  
In \cite{matas2018sim} the authors employed model free RL algorithms trained in simulation in an end-to-end manner  by resorting to expert demonstrations.
 Optimal expert demonstrations were also exploited in~\cite{Manocha_RAL2019} to derive a  controller based on random forests.

 In contrast, planning in a low-dimensional latent state space  significantly reduces the complexity of the input image space, albeit introducing the challenges for capturing the global structure and dynamics of the system in the latent space discussed in Sec.~\ref{sec:intro}.
Embed-to-Control  \cite{NIPS2015_e2c_watter} 
pioneers in learning 
a latent linear dynamical model for planning continuous actions. Variational inference was
used to infer a latent representation and dynamical system parameters to reconstruct a sequence of images.
  In addition to estimating transition and observation models,~\cite{hafner2018learning}
 proposed a deep planning network which also learns a reward function in the latent space.
The latter was then used to find viable trajectories resorting to a Model Predictive Control (MPC) approach.
A comparison between our method and  a baseline inspired by this approach can be found in Sec.~\ref{sec:sec:stack:lsr_performance}. 

 RL in the latent space was applied in~\cite{lynch2020learning}, 
where a VAE encodes trajectories into the latent space that is
optimized to minimize the KL-divergence between the proposed latent plans and those that have been encountered during self-play exploration.  Long-horizon visual planning was instead the focus of  \cite{pertch2020long}, which introduced latent space goal-conditioning to carry out long-horizon planning by reducing the search space and performing hierarchical optimization. 

The low dimensionality of the latent embeddings also enables the employment of traditional planning strategies 
in the latent space.   
In this regard, 
a framework for global search in a latent space  
 was designed in~\cite{Ichter2019} which is based on three components: \emph{i)} a latent state representation, \emph{ii)} a network to approximate the latent space dynamics, and \emph{iii)} a collision checking network. Motion planning is
then performed directly in the latent space by an RRT-based algorithm. 
In \cite{Ichter2021BroadlyExploringLT} the same authors combined the insights of RRT-based search in the latent space with the self play in \cite{lynch2020learning} and introduce Broadly-Exploring Local-policy Trees that produce long-horizon, sequential plans via a model-based, task-conditioned tree search.
 Imitation learning was instead leveraged in~\cite{srinivas2018universal}. In particular, a latent space Universal Planning Network 
was designed in~\cite{srinivas2018universal} to embed differentiable planning policies. The process is
learned in an end-to-end fashion from imitation learning and gradient descent is 
used to find optimal trajectories. Alternatively, a motion planning network with active learning procedure
was developed in~\cite{Yip_TRO2020} to reduce the data for training and actively ask for expert demonstrations only when needed.

 Graph structures have also been employed in the literature to perform planning in the latent space. In this regard, a graph neural network (GNN) %is
was  used in~\cite{kipf2019contrastive} to model the relations and transitions given the representations of objects in the scene, which were obtained with contrastive learning and Convolutional Neural Network (CNN).
 Moreover, combining RL with the idea of connecting states in the latent space via a graph 
was proposed in Semi-Parametric Topological Memory (SPTM) framework~\cite{savinov2018semiparametric}, where
an agent explores the environment and encodes observations into a latent space using a retrieval network. Each encoded observation forms a unique node in a memory graph built in the latent space. This graph is then used to plan an action sequence from a start to a goal observation using a locomotion network. 
As discussed in Sec.~\ref{sec:sec:stack:lsr_performance}, where we compare our method with the SPTM framework, the latter is
optimized for the continuous domain  with 
action/observation trajectories as input and 
 builds on the assumption that each observation is mapped to a unique latent code.
The work in~\cite{liu2020hallucinative} builds upon SPTM by additionally leveraging temporal closeness of the subsequent observations in the trajectories, while the study in~\cite{emmons2020sparse} performs merging of the same underlying states using  a two-way consistency criterion.

Latent representations are also suitable for tasks considering deformable objects as these are intrinsically hard to model analytically.
In \cite{Yan2020LearningPR}, contrastive learning 
was used 
to learn a predictive model in the latent space for planning rope and cloth flattening actions.
 In addition, \cite{Navarro_2021} proposed a feedback latent
representation framework for semantic soft object manipulation using geodesic path-based algorithms to perform planning in the latent space.

 In this work, we leverage weak labels extracted from demonstrated actions in the dataset
to capture the global structure of the state space and its dynamics in a data-efficient manner. More specifically, we build a graph  
in a low-dimensional latent state space to perform planning for rigid and deformable object manipulation tasks.

\section{Problem Statement and Notation} \label{sec:proplemdef}
{

\begin{table}[h!]
%\resizebox{\textwidth}{!}{%
\centering
\begin{tabularx}{\linewidth}{|l|X|}
\hline
Variable & Meaning  \\ \hline
$\mathcal{I}$ & Space of observations, \ie, images \\  \hline
$\mathcal{U}$ &  Space of actions  \\ \hline
$\mathcal{Z}$ &   Low-dimensional latent space  \\ \hline
$P_I,\,P_u,\,P_z$ &   Planned sequence of images, actions and latent states from assigned start and goal observations, respectively \\  \hline
$\mathcal{Z}_{sys}^i$ &  Covered region $i$ of the latent space \\ \hline
$\mathcal{Z}_{sys}$ & Overall covered region of the latent space \\ \hline
$\rho$ &  Specifics of the action that took place between two images $I_1$ and $I_2$\\ \hline
$\mathcal{T}_{I},\mathcal{T}_{z}$ &  Datasets containing image tuples ($I_1$,$I_2$,$\rho$) and their embeddings ($z_1$,$z_2$,$\rho$), respectively \\ \hline
$\xi$ &  Latent mapping function from $\mathcal{I}$ to $\mathcal{Z}$\\ \hline
$\omega$ & Observation generator function from $\mathcal{Z}$ to $\mathcal{I}$ \\ \hline
$d_m$ & Minimum distance encouraged among action pairs in the latent space \\ \hline
$p$ & Metric $L_p$ \\ \hline
$\tau$ & Clustering threshold for LSR building \\ \hline
$c_{\max}$ & Maximum number of connected components of the LSR \\ \hline
{
$N_{\varepsilon_z}(z)$} & {The $\varepsilon_z$-neighbourhood of a covered state $z$ containing same covered states} \\\hline
$\varepsilon^i$ &  $\epsilon_z$ associated with all the states in the covered region $\mathcal{Z}_{sys}^i$,  \ie $\varepsilon^i = \epsilon_z$  $\forall z \in \mathcal{Z}_{sys}^i$ \\\hline
\end{tabularx}%
%}
\caption{Main notations introduced in the paper.  }
\label{tab:notation}
\end{table}

 The goal of visual action planning, also referred to as \emph{``visual planning and acting"} in~\cite{wang2019learning},  can be formulated as follows: given start and goal images, generate a path as a sequence of images representing intermediate states and compute dynamically valid actions between them. 
We now formalize the problem and provide notation in Table~\ref{tab:notation}.

\begin{figure*}[htp]
\begin{center}
\includegraphics[width=\textwidth]{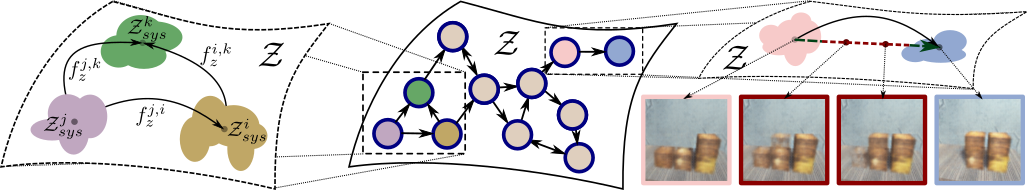}
\end{center}
\caption{Illustrative representation of the latent space $\mathcal{Z}$. In the middle, possible transitions (arrows) between covered regions (sketched with circles) are shown. On the left, details of the covered regions with different shapes and representative points are provided. On the right, observations from a box stacking tasks are shown. In detail, the ones obtained from covered regions (in pink and blue) contain meaningful task states, while the ones generated from not covered regions (in red) show fading boxes that do not represent possible states of the system. } 
\label{fig::visual_mani_plan}
\end{figure*}

Let $\mathcal{I}$ be the space of all possible observations of the system's states represented as images with fixed resolution and let $\mathcal{U}$ be the set of possible control inputs or actions.  

\begin{dfn}
A \emph{visual action plan} consists of a \emph{visual plan} represented as a sequence of images $P_I = \{I_{start} = I_{0}, I_{1} ...,  I_N = I_{goal}\}$ where $ I_{start}, I_{goal} \in \mathcal{I}$ are images  capturing the underlying start and goal states of the system, respectively, and an \emph{action plan} represented as a sequence of actions $P_u = \{u_0, u_1, ..., u_{N-1}\}$ where $u_n \in \mathcal{U}$ generates a transition between consecutive states contained in the observations $I_n$ and $I_{n+1}$ for each $n\in\{0, ..., N-1\}$. 
\end{dfn}

To retrieve the underlying states represented in the observations as well as to reduce the complexity of the problem we map $\mathcal{I}$ into a lower-dimensional latent space $\mathcal{Z}$ such that each observation $I_n \in \mathcal{I}$ is encoded as a point $z_n \in \mathcal{Z}$ extracting the state of the system captured in the image $I_n$. We call this map a  \emph{latent mapping} and denote it by $\xi: \mathcal{I} \to \mathcal{Z}$. In order to generate visual plans, we additionally assume the existence of a mapping $\omega: \mathcal{Z} \to \mathcal{I}$ called  \emph{observation generator}.

Let $\mathcal{T}_I = \{I_1, ..., I_{M}\} \subset \mathcal{I}$ be a finite set of input observations inducing a set of \emph{covered states} $\mathcal{T}_z = \{z_1, ..., z_M\} \subset \mathcal{Z}$, \ie, $ \mathcal{T}_z = \xi(\mathcal{T}_I)$. 
 In order to identify a set of unique covered states, we make the following assumption on $\mathcal{T}_z$. 

\begin{asm}
\label{eps-validity}
Let $z \in \mathcal{T}_{z}$ be a covered state. Then there exists $\varepsilon_z > 0$ such that any other state $z'$ in the $\varepsilon_z-$neighborhood $N_{\varepsilon_z}(z)$ of $z$ can be considered as the same underlying state. 
\end{asm} 

This allows both generating a valid visual action plan and taking into account the uncertainty induced by imprecisions in action execution. 
Let 
\begin{equation}
\label{eq:union-eps} \mathcal{Z}_{sys} = \bigcup_{z \in \mathcal{T}_z} N_{\varepsilon_z}(z) \subset \mathcal{Z}
\end{equation} 
be the union of $\varepsilon_z$-neighbourhoods of the covered states $z \in \mathcal{T}_z$. Given $\mathcal{Z}_{sys}$, a visual plan can be computed in the latent space using a \emph{latent plan} $P_z = \{z_{start} = z_0, z_1, ..., z_N = z_{goal}\}$, where $z_n \in \mathcal{Z}_{sys}$, which is then decoded with the observation generator $\omega$ into a sequence of images.

To obtain a valid visual plan, we study the structure of the space $\mathcal{Z}_{sys}$ which 
in general is not path-connected, i.e., does not contain all the points on linear interpolations between any two states $z_1, z_2 \in \mathcal{Z}_{sys}$. 
As we show in Fig.~\ref{fig::visual_mani_plan} on the right, such interpolation may result in a path containing points from $\mathcal{Z} - \mathcal{Z}_{sys}$ that do not correspond to covered states of the system  and are therefore not guaranteed to be meaningful.
To formalize this, we define an equivalence relation in $\mathcal{Z}_{sys}$
\begin{equation}\label{eq:equiv}
z \sim z' \iff \text{$z$ and $z'$ are path-connected in $\mathcal{Z}_{sys}$,}
\end{equation}
which induces a partition of the space $\mathcal{Z}_{sys}$ into $m$ equivalence classes $[z_1], \dots, [z_m]$. Each equivalence class $[z_i]$ represents a path-connected component of $\mathcal{Z}_{sys}$
\begin{equation}
\mathcal{Z}_{sys}^{i} = \bigcup_{z \in [z_i]} N_{\varepsilon_z} (z) \subset  \mathcal{Z}_{sys} \label{def:regions}
\end{equation}
called \textit{covered region}. 
To connect the covered regions, we define a set of transitions between them: 

\begin{dfn}
A \emph{transition function}  $f^{i, j}_z: \mathcal{Z}^{i}_{sys} \times \mathcal{U} \to \mathcal{Z}^{j}_{sys}$ maps any point $z \in \mathcal{Z}^{i}_{sys}$ to an equivalence class representative $z_{sys}^j \in \mathcal{Z}^{j}_{sys}$, where $i, j \in \{1, 2,..., m\}$ and $i \neq j$.
\end{dfn}

Equivalence relation~\eqref{eq:equiv} and Assumption~\ref{eps-validity} imply that two distinct observations $I_1$ and $I_2$ which are mapped into the same covered region $\mathcal{Z}^i_{sys}$ contain the same underlying state of the system, and can be represented by the same equivalence class representative $z_{sys}^i$. Given a set of covered regions $\mathcal{Z}_{sys}^i$ in $\mathcal{Z}_{sys}$ and a set of transition functions connecting them we can approximate the global transitions of $\mathcal{Z}_{sys}$ as shown in Fig.~\ref{fig::visual_mani_plan} on the left. To this end, we define a Latent Space Roadmap (see Fig.~\ref{fig::visual_mani_plan} in the middle):

\begin{dfn}
\label{def:lsr}
A Latent Space Roadmap is a directed graph $\LSR = (\mathcal{V}_{\LSR}, \mathcal{E}_{\LSR})$ where each vertex $v_i \in \mathcal{V}_{\LSR} \subset \mathcal{Z}_{sys}$ for $i \in \{1, 2,..., m\}$ is an equivalence class representative of the covered region $\mathcal{Z}^{i}_{sys}  \subset \mathcal{Z}_{sys}$, and an edge $e_{i, j} = (v_i, v_j) \in \mathcal{E}_{\LSR}$ represents a transition function $f_z^{i, j}$ between the corresponding covered regions $\mathcal{Z}^{i}_{sys}$ and $\mathcal{Z}^{j}_{sys}$ for $i \neq j$. Moreover, weakly connected components of an $\LSR$ are called \emph{graph-connected components.}
\end{dfn}

}

\section{Methodology}\label{sec:overview}
We first present the structure of the training dataset and then provide an overview of the approach.

% --- FIGURE: OVERVIEW --- %
\begin{figure*}[htb!]
\begin{center}
\includegraphics[width=0.7\textwidth]{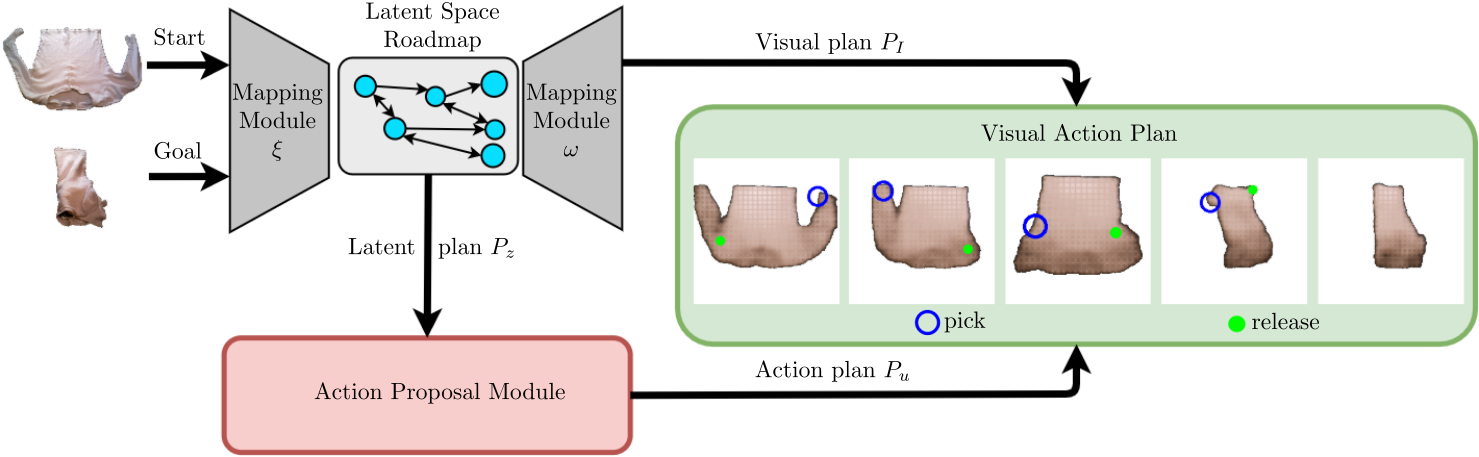}
\end{center}
\caption{Overview of the proposed method. Start and goal images (left) are mapped to the latent space $\mathcal{Z}$ by the latent mapping $\xi$. A latent plan is then found with the LSR (cyan circles and arrows) and is \textit{decoded} to a visual plan using the observation generator $\omega$. The APM (red) proposes actions to achieve the transitions between states in the visual plan. The final result is a \emph{visual action plan} (green) from start to goal. 
A re-planning step can also be added after every action to account for execution uncertainties as in  Fig.~\ref{fig:folding:super_results_folding}.
}
\label{fig::overview_pipeline}
\end{figure*}
% --- FIGURE: OVERVIEW --- %

\subsection{Training Dataset}\label{ssec:dataset}
\begin{figure}[htb!]
\begin{center}
\includegraphics[width=0.4\textwidth]{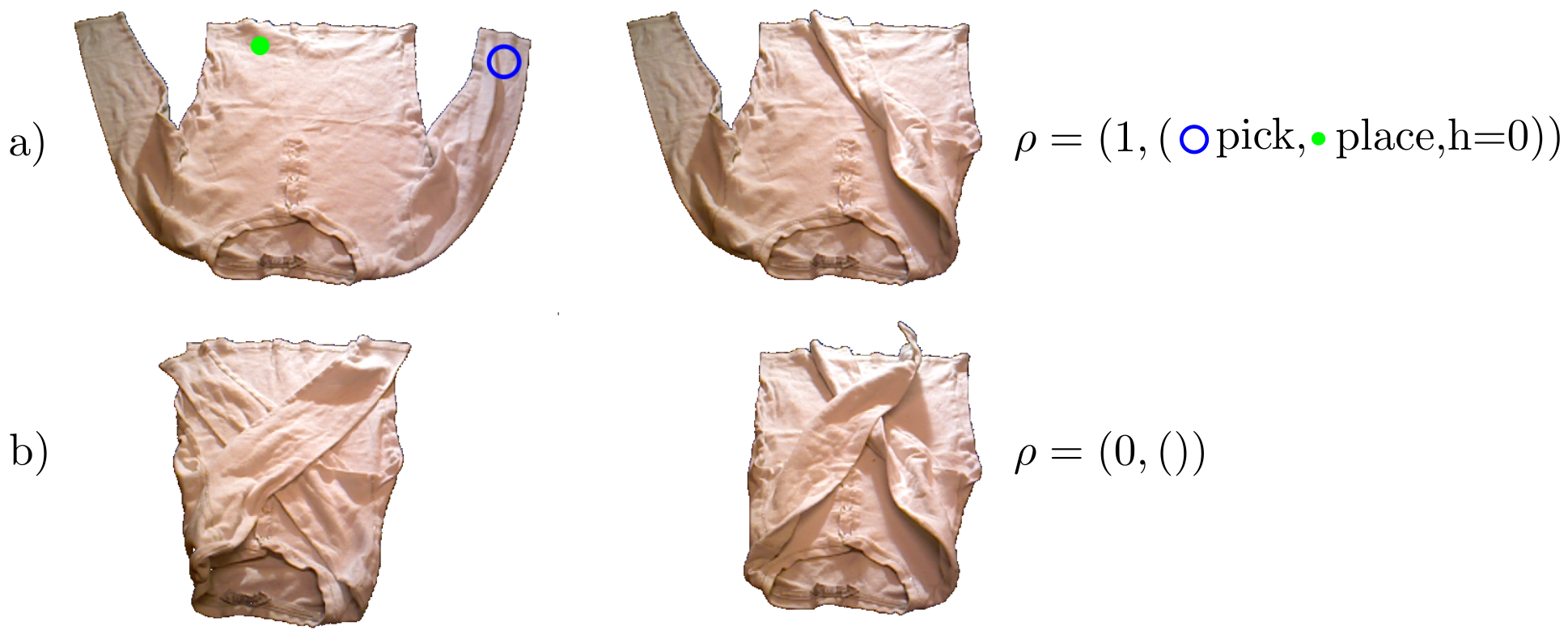}
\end{center}
\caption{Example of action (a) and no-action (b) pairs in folding task. 
}
\label{fig::act-no-ac-fold}
\end{figure}

We consider a training dataset $\mathcal{T}_I$ consisting of generic tuples of the form $(I_1, I_2, \rho)$ where $I_1 \subset \mathcal{I}$ is an image of the start state, $I_2 \subset \mathcal{I}$ an image of the successor state, and $\rho$ a variable representing the action that took place between the two observations. Here, an action is considered to be a \textit{single} transformation that produces any consecutive state represented in $I_2$ different from the start state in $I_1$, \ie, 
$\rho$ cannot be a composition of several transformations. 
On the contrary, we say that no action was performed if images $I_1$ and $I_2$ are observations of the same state, \ie, 
 if $\xi(I_1) \sim \xi(I_2)$ with respect to the equivalence relation~\eqref{eq:equiv}.
The variable $\rho = (a, u)$ consists of a binary variable $a \in \{0, 1\}$ indicating whether or not an action occurred as well as a variable $u$ containing the task-dependent action-specific information. The latter, if available, is used to infer the transition functions $f_z^{i, j}$. We call a tuple $(I_1, I_2, \rho = (1, u))$ an \textit{action pair} and  $(I_1, I_2, \rho = (0, u))$ a \textit{no-action pair}. For instance, Fig.~\ref{fig::act-no-ac-fold} shows an example of an action pair (top row) and a no-action pair (bottom row) for the folding task. In this case, the action specifics $u$ contain the pick and place coordinates to achieve the transition from the state  captured by $I_1$ to the state  captured by $I_2$, while the no-action pair images are  \emph{different} observations of the same underlying state of the system  represented by slight perturbations of the sleeves. When the specifics of an action $u$ are not needed, we omit them from the tuple notation and simply write $(I_1, I_2, a)$.
By abuse of notation, we sometimes refer to an observation $I$ contained in any of the training tuples as $I \in \mathcal{T}_I$. Finally, we denote by $\mathcal{T}_z$ the encoded training dataset $\mathcal{T}_I$ consisting of latent tuples $(z_1,z_2,\rho)$ obtained from the input tuples $(I_1,I_2,\rho) \in \mathcal{T}_I$ by encoding the inputs $I_1$ and $I_2$ into the latent space $\mathcal{Z}_{sys}$ with the latent mapping $\xi$. The obtained states $z_1, z_2 \in \mathcal{Z}_{sys}$ are called \textit{covered states}.

\begin{remark}
The dataset $\mathcal{T}_I$ is not required to contain all possible action pairs of the system but only a subset of them  that sufficiently cover the  dynamics, which makes our approach data efficient.
\end{remark}

\subsection{System Overview}
Generation of visual action plans consists of three components visualized in Fig.~\ref{fig::overview_pipeline}: 
\begin{itemize}
    \item \textbf{Mapping Module (MM)} used to both extract a low-dimensional representation of a state represented by a given observation, and to generate an  exemplary observation from a given latent state  (Sec.~\ref{sec:mapping_module}); 
    \item \textbf{Latent Space Roadmap (LSR)} built in the low dimensional latent space and used to plan  (Sec.~\ref{sec:lsr}); 
    \item \textbf{Action Proposal Module (APM)} used to predict action specifics for executing a latent plan found by the LSR (Sec.~\ref{sec:apm}). 
\end{itemize}

The MM consists of the latent mapping $\xi: \mathcal{I} \to \mathcal{Z}$ and the observation generator $\omega: \mathcal{Z} \to \mathcal{I}$. To find a visual plan between a given start observation $I_{start}$ and goal observation $I_{goal}$, the latent mapping $\xi$ first extracts the corresponding lower-dimensional representations $z_{start}$ and  $z_{goal}$ of the underlying start and goal states, respectively. Ideally, $\xi$ should perfectly extract the underlying state of the system such that different observations containing the same state are mapped into the same latent point. In practice, however, the unknown true latent embedding $\xi$ is \textit{approximated} with a neural network which implies that different observations containing the same state could be mapped to different latent points. In order to perform planning in $\mathcal{Z}$, we thus build the LSR which is a graph-based structure identifying the latent points belonging to the same underlying state and approximating the system dynamics. This enables finding the latent plans $P_z$ between the extracted states $z_{start}$ and  $z_{goal}$. For the sake of interpretability, latent plans $P_z$ are {\it decoded}
  into visual plans $P_I$, consisting of a sequence of images, by the observation generator $\omega$.  

 We complement the generated visual plan $P_I$ with the action plan $P_u$ produced by the APM, which proposes an action $u_i$ that achieves the desired transition \mbox{$f_z^{i, i+1}(z_i, u_i) = z_{i+1}$} between each pair $(z_{i}, z_{i+1})$ of consecutive states in the latent plan $P_z$ found by the LSR.

The visual action plan produced by the three components can be executed by any suitable framework. 

\begin{remark}\label{rem:replanning}
If open loop execution is not sufficient for the task, as for deformable object manipulation, a re-planning step can be added after every action. This implies that a new visual action plan is produced after the execution of each action until the goal is reached. A visualization of the re-planning procedure is shown in Fig.~\ref{fig:folding:super_results_folding} 
on the T-shirt folding task presented in Sec.~\ref{sec:exp}. 
\end{remark}

\begin{remark}
Our method is able to generate a sequence of actions $\{u_0, \dots, u_{N-1}\}$ to reach a goal state in $I_N$ from a given start state represented by $I_0$, even though the tuples in the input dataset $\mathcal{T}_I$  only contain \textit{single} actions $u$ that represent the weak supervision signals.
\end{remark}

\section{Mapping Module (MM)}
\label{sec:mapping_module}
The mappings $\xi: \mathcal{I} \to \mathcal{Z}$ and $\omega: \mathcal{Z} \to \mathcal{I}$ as well as the low-dimensional space $\mathcal{Z}$ can be realized using any encoder-decoder based algorithms, for example VAEs, AEs or Generative Adversarial Networks (GANs) combined with an encoder network.  The primary goal of MM is to find the best possible approximation $\xi$ such that the structure of the extracted states in the latent space $\mathcal{Z}$ resembles the one corresponding to the unknown underlying system. The secondary goal of MM is to learn an observation generator $\omega$ which enables visual interpretability of the latent plans. Since the quality of these depends on the structure of the latent space $\mathcal{Z}$,
we leverage the action information contained in the binary variable $a$ of the training tuples $(I_1, I_2, a)$  to improve the quality of the latent space. We achieve this by introducing a contrastive loss term~\cite{hadsell2006dimensionality} which can be easily added to the loss function of any algorithm used to model the MM.

More precisely, we introduce a general \emph{action} term
\begin{equation}\label{eq:acloss}
\mathcal{L}_{action}(I_1, I_2) \!=\! \begin{cases}
			\max(0, d_m - ||z_1-z_2||_p) & \text{if } a = 1\\
            ||z_1 - z_2||_p & \text{if } a = 0
		 \end{cases}
\end{equation}
where $z_1, z_2 \subset \mathcal{Z}_{sys}$ are the latent encodings of the input observations $I_1, I_2 \subset \mathcal{T}_I$, respectively,  $d_m$ is a hyperparameter, and the subscript $p \in \{1, 2, \infty\}$ denotes the metric $L_p$. The action term $\mathcal{L}_{action}$ naturally imposes the formulation of the covered regions $\mathcal{Z}_{sys}^i$ in the latent space. On one hand, it encodes identical states contained in the no-action pairs close by. On the other hand, it encourages different states to be encoded in separate parts of the latent space via the hyperparameter $d_m$.

\begin{figure}[htb!]
\begin{center}
\includegraphics[width=\linewidth]{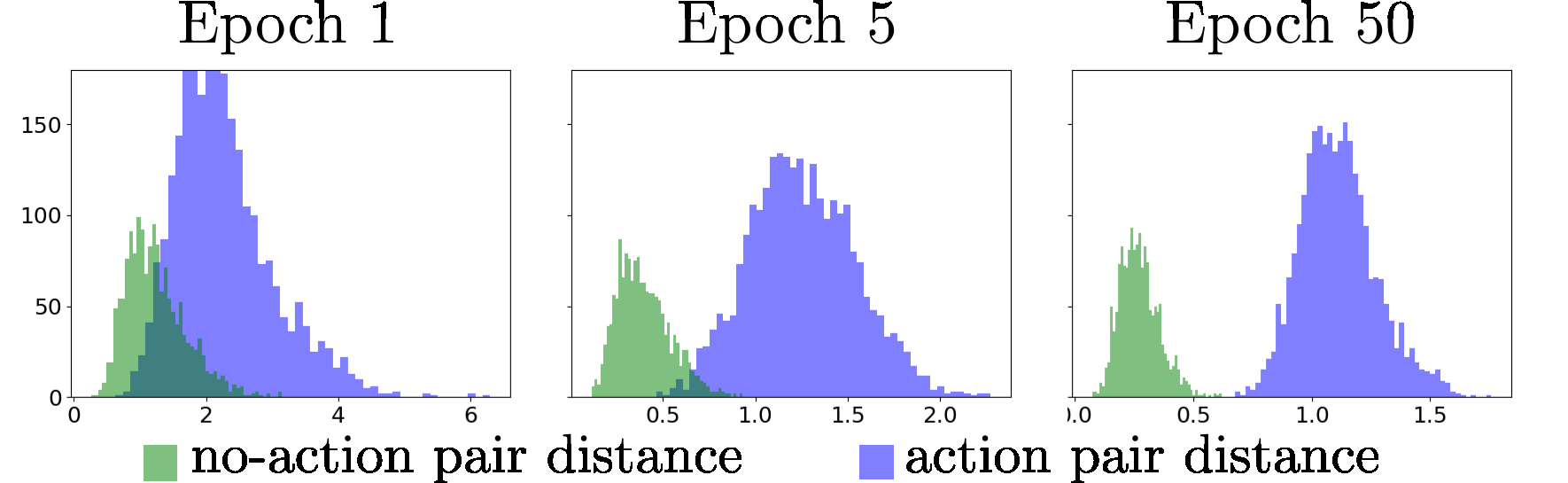}
\end{center}
\caption{An example showing histograms of distances $||z_1 - z_2||_1$ for latent action (in blue) and no-action pairs (in green) obtained at epochs $1$, $5$ and $50$ during the training of VAE on the hard box stacking task (more details in Sec.~\ref{sec:box}). The figure shows the separation of the action and no-action distances induced by dynamically  increasing  the minimum distance $d_m$ in $\mathcal{L}_{action}$.
}
\label{fig::d_m_histo}
\end{figure}

As we experimentally show in Sec.~\ref{sec:sec:stack:dm}, the choice of $d_m$ has a substantial impact on the latent space structure. Therefore, we propose to learn its value \textit{dynamically}  during the training of the MM.  In particular, $d_m$ is increased until the separation of action and no-action pairs is achieved. Starting from $0$ at the beginning of the training, we increase $d_m$  by $\Delta d_m$ every $k$th epoch as long as the maximum distance between no-action pairs is larger then the minimum distance between action pairs. The effect of dynamically increasing $d_m$ is shown in Fig.~\ref{fig::d_m_histo} where we visualize the distance  $||z_1 - z_2||_1$ %$||z_1 - z_2||_d$
between the latent encodings of every action training pair (in blue) and no-action training pair (in green) obtained at various epochs during training on a box stacking task. It can be clearly seen that the parameter $d_m$ is increased as long as there is an  intersection between action and no-action pairs.  Detailed investigation of this approach  as well as its positive effects on the structure of the latent space are provided in Sec.~\ref{sec:sec:stack:dm}. Note that the dynamic adaptation of the parameter $d_m$ eliminates the need to predetermine its value as in our previous work~\cite{ouriros}.

We use a VAE such that its latent space represents the space $\mathcal{Z}$, while the encoder and decoder networks realize the mappings $\xi$ and $\omega$, respectively. We validate  this choice in Sec.~\ref{sec:sec:stack:vae_vs_ae} by comparing it to AE. In the following, we first provide a brief summary of the VAE framework~\cite{kingma2013auto, rezende2014stochasticvae2} and then show how the action term can be integrated into its training objective.
Let $I \subset \mathcal{T}_I$ be an input image, and let $z$ denote the unobserved latent variable with prior distribution $p(z)$. The VAE model consists of encoder and decoder neural networks that are jointly optimized to represent the parameters of the approximate posterior distribution $q(z|I)$ and the likelihood function $p(I|z)$, respectively. In particular, VAE is trained to minimize 
\begin{align}
\mathcal{L}_{vae}(I) \!=\! E_{z \sim q(z|I)}[\log p(I|z)] + \beta \!\cdot\! D_{KL}\left(q(z|I) || p(z)\right) \label{eq:vaeloss}
\end{align} 
with respect to the parameters of the encoder and decoder neural networks. The first term influences the quality of the reconstructed samples, while the second term, called Kullback-Leibler (KL) divergence term, regulates the structure of the latent space. The trade-off between better reconstructions or a more structured latent space is controlled by the parameter $\beta$, where using a $\beta > 1$ favors the latter~\cite{higgins2016beta, burgess2018understanding}. The action term~\eqref{eq:acloss} can be easily added to the VAE loss~\eqref{eq:vaeloss} as follows:
\begin{equation}\label{eq:loss}
\mathcal{L}(I_1, I_2) =  \frac{1}{2}( \mathcal{L}_{vae}(I_1) +\mathcal{L}_{vae}(I_2))   + \gamma \cdot \mathcal{L}_{action}(I_1, I_2)
\end{equation}
where $I_1, I_2 \subset \mathcal{T}_I$ and the parameter $\gamma$ controls the influence of the distances among the latent encodings on the latent space structure.  Note that the same procedure applies for integrating the action term~\eqref{eq:acloss} into any other framework that models the MM. \\

\vspace{-10pt}

\section{Latent Space Roadmap (LSR)}\label{sec:lsr}
The Latent Space Roadmap, defined in \textit{Definition~\ref{def:lsr}}, is built in the latent space $\mathcal{Z}$ obtained from the MM. 
LSR is a graph that enables planning in the latent space which identifies sets of latent points associated with the same underlying state and viable transitions between them.
Each node in the roadmap is associated with a covered region $\mathcal{Z}_{sys}^i$. Two nodes are connected by an edge if there exists an action pair $(I_1, I_2, \rho = (1,u_1))$ in the training dataset $\mathcal{T}_I$ such that the transition $f_z^{1, 2}(z_1, u_1) = z_2$ is achieved in~$\mathcal{Z}_{sys}$. 

 The LSR building procedure is summarized in Algorithm \ref{alg::fix_epsilon} and discussed in Sec.~\ref{ssec:LSR_building}. It relies on a clustering algorithm that builds the LSR using the encoded training data $\mathcal{T}_z$ and a specified metric $L_p$ as inputs. The input parameter $\tau$ is inherited from the clustering algorithm and we automatically determine it using the procedure described in Sec.~\ref{ssec:LSR_optim}.

\begin{algorithm} \caption{LSR building}
\small
\def\negsp{\vspace{-5pt}}
\def\negup{\vspace{-7pt}}
\def\negin{\vspace{-3pt}}
\setstretch{1.2}
\begin{algorithmic}[section]
\Require Dataset $\mathcal{T}_z$, metric $L_p$, clustering threshold $\tau$
%Phase 1
\negin
\State Phase 1
\negsp
\begin{algorithmic}[1]
\State init graph $\mathcal{G} = (\mathcal{V}, \mathcal{E}) := (\{\}, \{\})$
\ForEach {($z_1,z_2, a) \in \mathcal{T}_z $}
    \State $\mathcal{V} \gets $ create nodes $z_1,z_2$
    \If{$a=1$}
        \State $\mathcal{E} \gets $ create edge ($z_1,z_2$)
    \EndIf
\EndFor
\end{algorithmic}
\negup
\end{algorithmic}
\begin{algorithmic}[section]
\State Phase 2 
\negsp

%Phase 2
\begin{algorithmic}[1]

\State ${M} \gets \text{Average-Agglomerative-Clustering}(\mathcal{T}_z, L_p)$~\cite{mullner2011modern} 
\State $\mathcal{W} \gets \text{get-Disjoint--Clusters}({M}, \tau)  $ 
\State $\mathcal{Z}_{sys} \gets \{\}$
\ForEach {$\mathcal{W}^{i} \in \mathcal{W} $}
    \State $\varepsilon^i \gets \text{get-Cluster-Epsilon}(\mathcal{W}^{i})$
    \State $\mathcal{Z}^{i}_{sys} := \cup_{w \in \mathcal{W}^{i}} N_{\varepsilon^i}(w) $
    \State $\mathcal{Z}_{sys} := \mathcal{Z}_{sys} \cup \{\mathcal{Z}^{i}_{sys}\}$ 

\EndFor

\end{algorithmic}

\end{algorithmic}

\begin{algorithmic}[section]
\negup
    \State Phase 3
    \negsp
    \begin{algorithmic}[1]
        %Phase 3
        \State init graph $\text{LSR} = (\mathcal{V}_{\LSR}, \mathcal{E}_{\LSR}) := (\{\}, \{\})$
        \ForEach {$\mathcal{Z}^{i}_{sys} \in \mathcal{Z}_{sys}$}
            \State $w^i := \frac{1}{|\mathcal{W}^{i}|} \sum_{w \in \mathcal{W}^{i}}w$ 
            \State $z^i_{sys} := \text{argmin}_{z \in \mathcal{Z}_{sys}^i} ||z - w^i||_p$
            \State $\mathcal{V}_{\LSR} \gets $ create node $z^i_{sys}$
        \EndFor
        \negsp
        \ForEach {edge $ e = (v_1, v_2) \in \mathcal{E} $}
            \State find $\mathcal{Z}^{i}_{sys}, \mathcal{Z}^{j}_{sys}$ containing $v_1, v_2$, respectively 
            \State $\mathcal{E}_{\LSR} \gets $ create edge ($z^{i}_{sys},z^{j}_{sys}$)
        \EndFor
    \end{algorithmic}
    \negsp
    \Return LSR
    
\end{algorithmic}
\label{alg::fix_epsilon}
\end{algorithm}

\subsection{LSR Building} \label{ssec:LSR_building}
\begin{figure*}[htb!]
\begin{center}
\includegraphics[width=\linewidth]{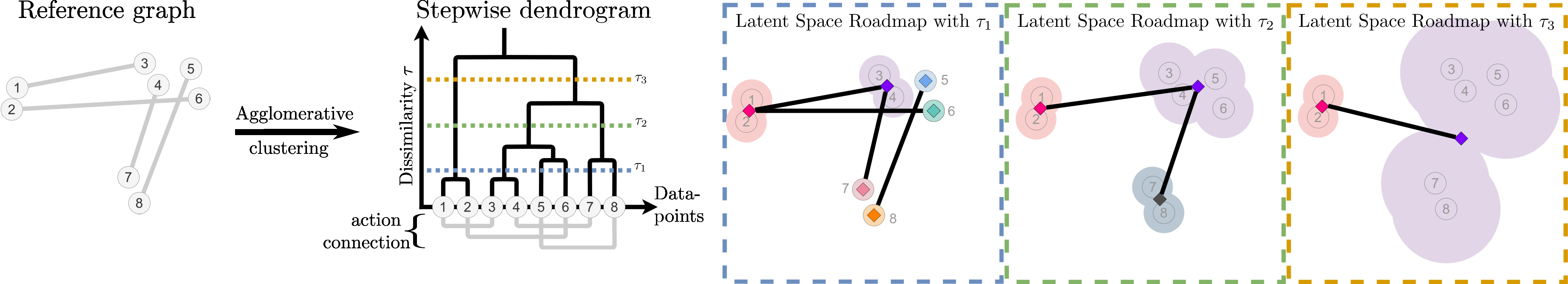}
\end{center}
\caption{Illustrative example visualising the LSR building steps and the effect of the clustering threshold $\tau$. The left shows the reference graph built in Phase $1$ of Algorithm \ref{alg::fix_epsilon}. The middle part visualizes a dendrogram $M$ obtained from the clustering algorithm in Phase $2$. On the right, three examples of LSRs are shown together with the covered regions (marked with various colors) corresponding to different clustering thresholds $\tau$ (with $\tau_1<\tau_2<\tau_3$) chosen from $M$. 
}
\label{fig:lsr:p2_explained}
\end{figure*}

Algorithm \ref{alg::fix_epsilon} consists of three phases. In Phase $1$ (lines $1.1 - 1.5$), we build a \emph{reference} graph $\mathcal{G} = (\mathcal{V}, \mathcal{E})$ induced by $\mathcal{T}_z$ and visualized on the left of Fig.~\ref{fig:lsr:p2_explained}. Its set of vertices $\mathcal{V}$ is the set of all the latent states in $\mathcal{T}_z$, while edges exist only among the latent action pairs. It serves as a look-up graph to  preserve the edges that later induce the transition functions~$f_z^{i, j}$. 

In Phase $2$, Algorithm \ref{alg::fix_epsilon} identifies the covered regions $\mathcal{Z}_{sys}^i \subset \mathcal{Z}_{sys}$. 
We achieve this by first 
clustering the training samples and then retrieving the covered regions from these clusters. 
We start by performing agglomerative clustering~\cite{mullner2011modern} on the encoded dataset $\mathcal{T}_z$ (line 2.1). Agglomerative clustering is a hierarchical clustering scheme that starts from single nodes of the dataset and merges the closest nodes, according to a dissimilarity measure, step by step until only one node remains. It results in a \textit{stepwise dendrogram} $M$, depicted in the middle part of Fig.~\ref{fig:lsr:p2_explained}, which is a tree structure visualizing the arrangement of data points in clusters with respect to the level of dissimilarity between them. We choose to measure this inter-cluster dissimilarity using the \textit{unweighted average} distance between points in each cluster, a method also referred to as UPGMA~\cite{sokal1958statistical}. More details about other possible clustering algorithms and dissimilarity measures are discussed in Sec.~\ref{sec:sec:stack:lsr_phase2}. 
Next, the dissimilarity value $\tau$, referred to as \textit{clustering threshold}, induces the set of disjoint clusters $\mathcal{W}$, also called \emph{flat} or \emph{partitional} clusters~\cite{celebi2014partitional}, from the stepwise dendrogram $M$~\cite{mullner2011modern} (line $2.2$). 
Points in each cluster $\mathcal{W}^i$ are then assigned a uniform $\epsilon^i$ (line $2.5$), \ie the neighbourhood size from Assumption~\ref{eps-validity} of each point $z\in\mathcal{W}^i$ is $\epsilon_z = \epsilon^i$. We discuss the definition of the $\epsilon^i$ value at the end of this phase. 
The union of the $\epsilon^i$-neighbourhoods of the points in $\mathcal{W}^i$ then forms the covered region $\mathcal{Z}^i_{sys}$ (line $2.6$). Illustrative examples of covered regions obtained from different values of $\tau$ are visualized on the right of Fig.~\ref{fig:lsr:p2_explained} using various colors. The optimization of $\tau$ is discussed in Appendix\ref{sec:app:clus-ae}.
The result of this phase is the set of the identified covered regions $\mathcal{Z}_{sys} = \{\mathcal{Z}_{sys}^i\}$ (line $2.7$). 

We propose to approximate $\epsilon^i$
as
\begin{equation}\label{eq:cluster_eps}
\epsilon^i = \mu^i  + \sigma^i
\end{equation}
where $\mu^i$ and $\sigma^i$ are the mean and the standard deviation of the distances $\|z_{j}^i - z_{k}^i\|_p$ among all the training pairs $(z_{j}^i,z_{k}^i) \in \mathcal{T}_z$ belonging to the $i$th cluster.
The approximation in \eqref{eq:cluster_eps} allows to take into account the cluster density such that denser clusters get lower $\epsilon^i$. 
In contrast to our previous work~\cite{ouriros}, we now enable clusters to have different $\epsilon$ values.  We validate the approximation~\eqref{eq:cluster_eps} in Secs.~\ref{sec:sec:stack:lsr_cover} and \ref{sec:sec:fold:lsr_cover} where we analyze the covered regions identified by the LSR. 

In Phase $3$, we build the $\text{LSR}= (\mathcal{V}_{\LSR}, \mathcal{E}_{\LSR})$. We first compute the mean value $w^i$ of all the points in each cluster $\mathcal{W}^i$ (line $3.3$). 
As the mean itself might not be contained in the corresponding path-connected component, we find the equivalence class representative $z_{sys}^i \in \mathcal{Z}_{sys}^i$ that is the closest (line $3.4$). The found representative then defines a node $v_i \in \mathcal{V}_{\LSR}$ representing the covered region $\mathcal{Z}_{sys}^i$ (line $3.5$). Lastly, we use the set of edges $\mathcal{E}$ in the reference graph built in Phase $1$ to infer the transitions $f_z^{i, j}$ between the covered regions identified in Phase $2$. We create an edge in LSR if there exists an edge in $\mathcal{E}$ between two vertices in $\mathcal{V}$ that were allocated to different covered regions (lines $3.6 - 3.8$). 
The right side of Fig.~\ref{fig:lsr:p2_explained} shows 
the final LSRs, obtained with different values of the clustering threshold $\tau$.

Note that, as in the case of the VAE (Sec.~\ref{sec:mapping_module}), no action-specific information $u$ is used in Algorithm~\ref{alg::fix_epsilon} but solely the binary variable $a$ indicating the occurrence of an action.

\subsection{Optimization of LSR Clustering Threshold $\tau$} \label{ssec:LSR_optim} 

The clustering threshold $\tau$, introduced in Phase $2$ of Algorithm \ref{alg::fix_epsilon}, heavily influences the number and form of the resulting clusters. Since there is no inherent way to prefer one cluster configuration over another, finding its optimal value is a non-trivial problem and subject to ongoing research ~\cite{langfelder2008defining}, \cite{bruzzese2015despota}, \cite{pasini2019adaptive}.
 However, in our case, since the choice of $\tau$ subsequently influences the resulting LSR,  we can leverage the information about the latter  to optimize  $\tau$. 
As illustrated in Fig.~\ref{fig:lsr:p2_explained}, the number of vertices and edges in $\text{LSR}_{\tau_i}$ changes with the choice of $\tau_i$. Moreover, the resulting LSRs can have different number of \textit{graph}-connected components. For example, $\text{LSR}_{\tau_1}$ in Fig.~\ref{fig:lsr:p2_explained} has $2$ graph-connected components, while $\text{LSR}_{\tau_2}$ and $\text{LSR}_{\tau_3}$ have only a single one. Ideally, we want to obtain a graph that exhibits both good connectivity which best approximates the true underlying dynamics of the system, and has a limited number of graph-connected component.
Intuitively, high number of edges increases the possibility to find latent paths from start to goal state. At the same time, this possibility is decreased when the graph is fragmented into several isolated components, which is why we are also interested in limiting the maximum number of graph-connected components.

While we cannot analyze the clusters themselves, we can  evaluate information captured by the LSR that correlates with the performance of the task, \ie, we can assess a graph by the number of edges and graph-connected components it exhibits as discussed above. This induces an objective which we can use to optimize the value of the clustering threshold $\tau$. We formulate it as  
\begin{equation}
    \psi(\tau, c_{\max}) =
    \begin{cases}
    |\mathcal{E}_{\LSR_\tau}| & \text{if } c_{\LSR_\tau}\leq c_{\max}, \\
    -\infty              & \text{otherwise},
\end{cases} \label{eq:brets_objective}
\end{equation}
where $|\mathcal{E}_{\LSR_\tau}|$ is the cardinality of the set $\mathcal{E}_{\LSR_\tau}$,  $c_{\LSR_\tau}$ represents the number of graph-connected components of the graph $\LSR_\tau$ induced by $\tau$, and the hyperparameter $c_{\max}$ represents the upper bound on the number of graph-connected components. 
The optimal $\tau$ in a given interval $[\tau_{\min}, \tau_{\max}]$ can be found by any scalar optimization method. 
In this work, we use 
Brent's optimization method~\cite{brent1971algorithm} maximizing~\eqref{eq:brets_objective}:
\begin{equation}\label{eq:brent_opt}
    \max_{\tau_{\min} \leq \tau \leq \tau_{\max}} \psi(\tau, c_{\max}).
\end{equation}
This optimization procedure is summarized in Algorithm \ref{alg::dendrogram_cut}. It takes as an input the encoded training data $\mathcal{T}_z$, the metric $L_p$, the search interval where the clustering parameter $\tau$ is to be optimized, and the upper bound $c_{\max}$ to compute the optimization objective in~\eqref{eq:brets_objective}. 
After initialization of the parameter $\tau$ (line $1$), for example, by considering the average value of its range, the Brent's optimization loop is performed (lines $2$-$5$). 
Firstly, the LSR with the current $\tau$ is built according to Algorithm~\ref{alg::fix_epsilon} (line $3$). Secondly, the optimization objective~\eqref{eq:brets_objective} is computed on the obtained $\LSR_\tau$ (line $4$). Thirdly, the parameter $\tau$ as well as the bounds $\tau_{\min}$ and $\tau_{\max}$ are updated according to~\cite{brent1971algorithm} (line $5$). The optimization loop is performed until the convergence is reached, \ie, until $|\tau_{\max}-\tau_{\min}|$ is small enough according to~\cite{brent1971algorithm}. Lastly,  the optimal $\tau^*$ (line $6$) is selected for the final $\LSR_{\tau^*}$.

Note that even though Algorithm~\ref{alg::dendrogram_cut} still needs the selection of the hyperparameter $c_{\max}$,
we show 
in Sec. \ref{sec:sec:stack:lsr_cmax} that it is rather robust to the choice of this parameter.

\begin{algorithm} \caption{LSR input optimization}
\small
\def\negsp{\vspace{-5pt}}
\def\negup{\vspace{-7pt}}
\def\negin{\vspace{-3pt}}
\setstretch{1.2}
\begin{algorithmic}[1]
\Require Dataset $\mathcal{T}_z$,  metric $L_p$, search interval $[\tau_{\min}, \tau_{\max}]$, $c_{\max}$
\State  $\tau \gets \text{init}(\tau_{\min}$, $\tau_{\max})$
\While{$|\tau_{\max} - \tau_{\min}| \text{ not small enough}$ }
    \State $\LSR_\tau \gets \text{LSR-building}(\mathcal{T}_z,  L_p, \tau) \quad$ [Algorithm \ref{alg::fix_epsilon}]
    \State $\psi \gets \text{Evaluate}(\LSR_\tau)\quad$ [Eq. \eqref{eq:brets_objective}]
    \State $\tau, \tau_{\min}, \tau_{\max} \gets \text{Brent-update} (\psi)\quad $~\cite{brent1971algorithm}  
\EndWhile
\State $\tau^* \gets \tau$ 
\end{algorithmic}
\Return $\LSR_{\tau^*}$

\label{alg::dendrogram_cut}
\end{algorithm}

%***********************************************
\subsection{Visual plan generation}
Given a start and goal observation, a trained VAE model and an LSR,
 the observations are first encoded by $\xi$ into the VAE's latent space $\mathcal{Z}$ where their closest nodes in the LSR are found. Next, all shortest paths
 in the LSR between the identified nodes are retrieved. 
Finally, the equivalence class representatives of the nodes comprising each of the found shortest path compose the respective latent plan $P_z$, which is then decoded into the visual plan $P_I$ using~$\omega$.

\section{Action Proposal Module (APM)}
\label{sec:apm}
 The final component of our framework is the Action Proposal Module (APM) which is used to complement a latent plan, produced by the LSR, with an action plan that can be executed by a suitable framework. The APM allows to generate the action plans from the extracted low-dimensional state representations rather than high-dimensional observations. The action plan $P_u$ corresponding to a latent plan $P_z$ produced by the LSR is generated sequentially: given two distinct consecutive latent states $(z_i, z_{i+1})$ from $P_z$, APM predicts an action $u_i$ that achieves the transition $f^{i, i+1}(z_i, u_i) = z_{i+1}$. Such functionality can be realized by any method that is suitable to model the action specifics of the task at hand.

We model the action specifics with a neural network called Action Proposal Network (APN). We deploy a  
multi layer perceptron and train it in a supervised fashion on the latent \emph{action} pairs 
obtained from the enlarged dataset $\mathcal{T}_z$ as described below.  We validate this choice in Sec~\ref{sec:exe:apm} where we compare it to different alternatives that produce action plans either by exploiting the LSR or by using the observations as inputs rather than extracted low-dimensional states.

The training dataset $\overline{\mathcal{T}_z}$ for the APN is derived from $\mathcal{T}_I$ but preprocessed with the VAE encoder representing the latent mapping $\xi$. We encode each training \emph{action} pair $(I_1, I_2, \rho = (1, u)) \in \mathcal{T}_I$ into $\mathcal{Z}$ and obtain the parameters $\mu_{i}, \sigma_{i}$ of the approximate posterior distributions $q(z|I_i) =  N(\mu_i, \sigma_i)$, for $i = 1, 2$. We then sample $2S$ novel points $z_1^s \sim q(z|I_1)$ and $z_2^s \sim q(z|I_2)$ for $s \in \{0, 1, \dots, S\}$. This results in $S+1$ tuples $(\mu_{1}, \mu_{2}, \rho)$ and $(z_1^s, z^s_2, \rho), 0 \le s \le S$, where $\rho = (1, u)$ was omitted from the notation for simplicity. The set of all such low-dimensional tuples forms the APN training dataset $\overline{\mathcal{T}_z}$. 

\begin{remark}
It is worth remarking the two-fold benefit of this preprocessing step: not only does it reduce the dimensionality of the APN training data but also enables enlarging it with novel points by factor $S+1$. Note that the latter procedure is not possible with non-probabilistic realizations of $\xi$.
\end{remark}

\section{Assumptions, Applicability and limitations of the method}\label{sec:limit}

In this section, we briefly overview our assumptions, describe tasks where our method is applicable, and  discuss its limitations.  
In order for our method to successfully perform a given visual action planning task, the observations contained in the training dataset $\mathcal{T}_I$ 
should induce the \textit{covered} states (defined in Sec.~\ref{sec:proplemdef}) that are considered in the planning. Furthermore, it is required that sufficiently many transitions among them are observed such that the obtained LSR adequately approximates the true underlying system dynamics. For example, the training datasets $\mathcal{T}_I$ in the box stacking tasks consist of $2500$ pairs of states of the system instead of all (i.e., $41616$) possible combinations. 
On the other hand, if the system contains many feasible states, it can be challenging to collect a dataset $\mathcal{T}_I$ that covers sufficiently many states  and transitions between them. 
Even though the performance of the LSR would deteriorate with such incomplete dataset, we do not consider this as the limitation of the method itself as this can be mitigated with online learning approaches, \eg, \cite{maeda2017active}, that dynamically adapt the LSR based on the interaction with the environment.

Given the assumptions on the format of the dataset $\mathcal{T}_I$ introduced in Sec.~\ref{ssec:dataset}, our method is best applicable to visual action planning tasks where feasible states of the system are finite and can be distinguished in $\mathcal{T}_I$ such that meaningful unambiguous actions to transition among them can be defined. 

Therefore, our approach does not generalize well to \textit{entirely novel} states of the system not contained in the training set. This is expected, as the model has no prior knowledge about the newly appeared state, such as, for example, an entirely new fold of a T-shirt or a new piece of garment. Such generalization could be achieved by integrating active learning approaches which is indeed an interesting future direction. 
We emphasise that the proposed method is not limited by the dimensionality of the system's states since that is reduced via MM. 
% --- Box Stacking --- %
\section{Simulation results}\label{sec:box}
\begin{figure}[htb!]
\begin{center}
\includegraphics[width=\linewidth]{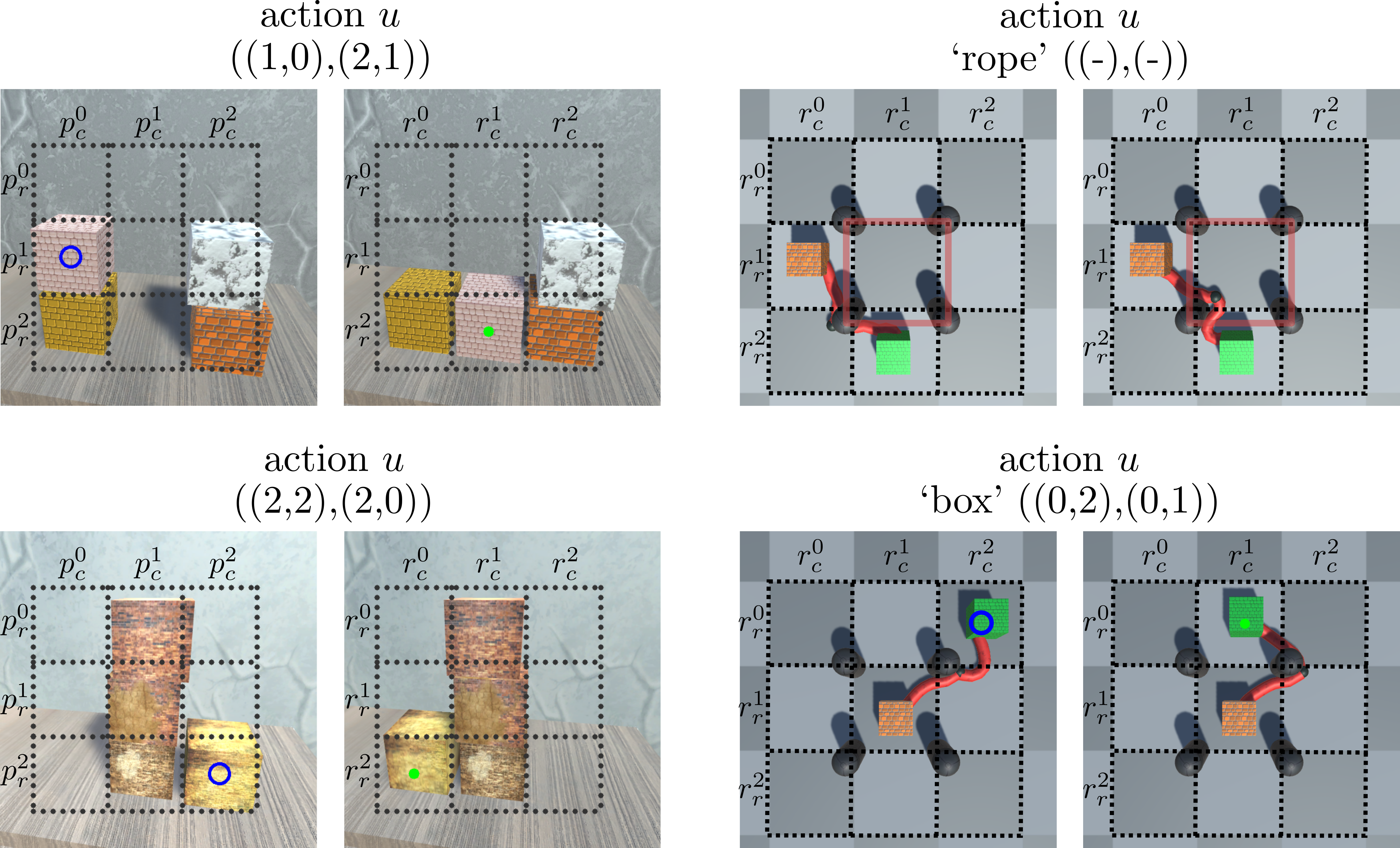}
\end{center}
\caption{Examples of actions $u$ in the normal (top) and hard (bottom) box stacking tasks (left) and  in the rope-box task (right). The blue circle shows the picking location $p$, and the green one the release position $r$. The action `rope' for moving the rope over the closest pillar is shown in top right. 
}
\label{fig:sim1:setup}
\end{figure}

 We experimentally evaluated our method on three different simulated tasks: two versions of a box stacking task (Fig.~\ref{fig:sim1:setup} left) and a  combined rope and box manipulation task (Fig.~\ref{fig:sim1:setup} right), which we refer to as \textit{rope-box manipulation} task.
 We considered the initial box stacking task used in our previous work \cite{ouriros} (top left), and a modified one where we made the task of retrieving the underlying state of the system harder. We achieved this by {\emph{i})} using more similar box textures which  made it more difficult to \emph{separate} the states, and \emph{ii}) by 
introducing different lighting conditions which made observations containing the same states look more dissimilar. We refer to the original setup as the \textit{normal stacking} task denoted by $ns$, and to the modified one as the \textit{hard stacking} task denoted by~$hs$. 
In the rope-box manipulation task  (Fig.~\ref{fig:sim1:setup} right), denoted by $rb$, a rope connects two boxes constraining their movement. To challenge the visual action planning, we again introduced different lighting conditions as well as the deformability of the rope. 

These three  setups enable automatic evaluation of the structure of the latent space $\mathcal{Z}_{sys}$, the quality of visual plans $P_I$ generated by the LSR and MM, and the quality of action plans ${P}_u$ predicted by the APN.  Moreover, they enable to perform a more thorough ablation studies on the introduced improvements of our framework which were not possible in our earlier version of the LSR~\cite{ouriros} since the resulting visual action plans achieved a perfect evaluation score.

All setups were developed with the Unity engine~\cite{unitygameengine} and the resulting images have dimension $256 \times 256 \times 3$. 
In the stacking tasks, four boxes with different textures that can be stacked in a $3 \times 3$ grid (dotted lines in Fig.~\ref{fig:sim1:setup}). A grid cell can be occupied by only one box at a time which can be moved according to the 
\emph{stacking rules}: i) it can be picked only if there is no other box on top of it, and ii) it can be released only on the ground or on top of another box inside the $3 \times 3$ grid. In both versions of the stacking task, the position of each box in a grid cell was generated by introducing $\sim 17\%$ noise along $x$ and $y$ axes which was applied when generating both action and no-action pairs. 
The action-specific information $u$, shown in Fig.~\ref{fig:sim1:setup} left,  is a pair $u = (p, r)$ of pick $p$ and release $r$ coordinates in the grid  modelled by the row and column indices, \ie, $p = (p_r, p_c)$ with $p_r, p_c \in \{0, 1, 2\}$, and equivalently for \mbox{$r = (r_r, r_c)$}.

In the rope-box manipulation task, two boxes and a rope  can be moved in a $3 \times 3$ grid  with $4$ pillars according to  the following  manipulation rules: i) a box can only be pushed one cell in the four cardinal directions but not outside the grid, ii) the rope can be lifted over the closest pillar, iii) the rope cannot be stretched over more that two cells, meaning the boxes can never be more than one move apart from being adjacent. In this task, the action-specific information $u$, shown in Fig.~\ref{fig:sim1:setup} right, denotes whether the rope is moved  over the closest pillar (top) or a box is moved in the grid (bottom) with respective pick $p$ and release $r$ coordinates.

% --- Figure Example box stacking --- %
% ---  %

According to the above rules, the training datasets $\mathcal{T}_I$ for stacking tasks contain all possible $288$ different grid configurations, \ie, the specification of which box, if any, is contained in each cell. In case of the rope-box manipulation task, $\mathcal{T}_I$ contains $157$ different grid configurations comprising the position of the rope and boxes. 
These $288$/$157$ grid configurations represent the covered states in these tasks. Note that the exact number of underlying states is in general not known. 
Given a pair of states and the task rules, it is possible to analytically determine whether or not an action is allowed between them. 
In addition, we can determine the grid configuration associated with an image (\ie, its underlying state) contained in the produced visual plan $P_I$ using classifiers. These were trained on the decoded images and achieved accuracy greater than $98.8\%$  on a holdout dataset composed of $750$ 
samples for both versions of the stacking task  and the rope-box task. All the implementation details can be found on our code repository\footnote{\label{fn:git} \url{https://github.com/visual-action-planning/lsr-v2-code}}. 

% -------------------- OBJECTIVES ------------------------%%%
\subsection{Experiment Objectives and Implementation Details}\label{sec:box-intro}

Our experiments are designed to answer the following questions: 
\begin{enumerate}
    \item \label{en:sim:mm}\textbf{MM} What is the impact of the action term~\eqref{eq:acloss} in the augmented loss function~\eqref{eq:loss} on the structure of the latent space? How do the respective parameters (\eg,  minimum distance) influence the overall $\LSR$ performance? Lastly, 
    how does the VAE framework perform compared to the AE one for modelling the mappings $\xi$ and $\omega$ in the MM? 
    \item \label{en:sim:lsr}\textbf{LSR}  What is the performance of the LSR compared to state of the art solutions like~\cite{savinov2018semiparametric} and~\cite{hafner2018learning}, and what is the influence of the action term~\eqref{eq:acloss} on it?  How do the respective LSR parameters (\eg, number of components) and the choice of the clustering algorithm impact the overall LSR performance?  How good is the LSR approximation of the covered regions? 
    \item \label{en:sim:apm}\textbf{APM} What is the performance of the APN model? 
\end{enumerate}

In this section, we present the implementation details and introduce the notation used to easily refer to the models in consideration. 
For VAEs (used in MM), each model is annotated by $\vae{ld}{task}{d}$ where $ld$ denotes the dimension of the latent space, $task$ denotes the version of the task and is either $ns$, $hs$ or $rb$ for the normal stacking task, hard stacking tasks or rope-box manipulation task, respectively. The parameter $d$ indicates whether or not the model was trained with the action loss term~\eqref{eq:acloss}. We use $d = b$ to denote a \emph{baseline} VAE trained with the original VAE objective~\eqref{eq:vaeloss}, and $d = L_1$ to denote an \emph{action} VAE trained with the loss function~\eqref{eq:loss} including the action term~\eqref{eq:acloss} using metric $L_1$. 
 Compared to~\cite{ouriros}, we consider only $L_1$ metric in our simulated experiments due to its superior performance over the $L_2$ and $L_{\infty}$ metrics established in~\cite{ouriros}.

All VAE models used a ResNet architecture~\cite{ResNet} for the encoder and decoder networks.
They were trained for $500$ epochs on a training dataset  $\mathcal{T}_I$, composed of $65\%$ action pairs and $35\%$ no-action pairs for stacking tasks, and $50\%$ action pairs and $50\%$ no-action pairs for rope-box manipulation task. For each combination of parameters ${ld}$, $task$, and $d$, we trained $5$ VAEs initialized with different random seeds. Same seeds were also used to create training and validations splits of the training dataset. The weight $\beta$ in~\eqref{eq:vaeloss} and~\eqref{eq:loss} was gradually increased from $0$ to $2$ over $400$ epochs,  
while $\gamma$ was fixed to $100$. In this way,  models were encouraged to first learn to reconstruct the input images and then to gradually structure the latent space. 
The minimum distance $d_m$ was dynamically increased every fifth epoch starting from $0$ using $\Delta d_m=0.1$ as described in Sec.~\ref{sec:mapping_module}. The effect of this dynamic parameter increase is shown in Fig.~\ref{fig::d_m_histo}. 

For LSR, we denote by $\lsr{1}$ a graph built using the metric $L_1$ in Algorithm~\ref{alg::fix_epsilon}. The parameters $\tau_{\min}$ and $\tau_{\max}$ in the LSR optimization~\eqref{eq:brent_opt} were set to $0$ and $3$, respectively. Unless otherwise specified, we fixed $ld = 12$ 
for all tasks. 
Moreover, the number of graph-components $c_{\max}$ in the optimization of the clustering threshold~\eqref{eq:brets_objective} was set to $1$ for $ns$, and $20$ for $hs$ and $rb$. These choices are explained in detail in the following sections.
Given an LSR, we evaluated its performance by measuring the quality of the visual plans found between $1000$ randomly selected start and goal observations from an unseen holdout set containing $2500$ images.  To automatically check the validity of the found paths, we 
used the classifiers on the observations contained in the visual plans to get the respective underlying states. We then defined a checking function (available on the code repository) that, given the states in the paths, determines whether they are allowed or not according to the the stacking or the manipulation rules.
In the evaluation of the planning performance we considered the following quantities: \emph{i)} percentage of cases when all shortest paths from start to goal observations are correct, denoted as \textit{\% All},  \emph{ii)}  percentage  of cases when at least one of the proposed paths is correct,  denoted as \textit{\% Any}, and \emph{iii)} percentage of correct single transitions in the paths,  denoted as \textit{\% Trans}. We refer to the \% Any score in \emph{ii)} as \textit{partial scoring}, and to the combination of scores \emph{i)}-\emph{iii)} as \textit{full scoring}. 
Mean and standard deviation values are reported over the $5$ different random seeds used to train the VAEs. 

For APNs, we use the notation $\apn{ld}{task}{d}$ analogous to the VAEs. The APN models were trained for $500$ epochs on the training dataset $\overline{\mathcal{T}_z}$ obtained following the procedure described in Sec~\ref{sec:apm} using $S = 1$. 
Similarly as for LSR, we report the mean and standard deviation of the performance obtained over the $5$  random seeds used in the VAE training.

%---------------------------------------------------------------
%----------------------------------------------------------------
%-----------------------------------------------------------------------------------------------------------------
\subsection{MM Analysis}\label{sec:sim:mm}

In this section, we validate the MM module answering the questions in point~\ref{en:sim:mm}) of Sec.~\ref{sec:box-intro}. 
In the first experiment, we investigated the influence of the dynamic parameter $d_m$ on the LSR performance. We then studied the structure of the latent space by analyzing the distance between  encodings of different states. 
Lastly, we compared the LSR performance when modelling MM with an AE framework instead of a VAE.

\subsubsection{Influence of dynamic $\boldsymbol{d_m}$}
\label{sec:sec:stack:dm}
A key parameter in the action term~\eqref{eq:acloss} is the minimum distance $d_m$ encouraged among the action pairs. We considered the hard stacking  and rope-box manipulation tasks and validated the approach proposed in Sec.~\ref{sec:mapping_module}, which dynamically increases $d_m$ to separate action and no-action pairs (see Fig.~\ref{fig::d_m_histo}). At the end of the training, the approach results in $d_m =2.3 \pm 0.1$ and $d_m=2.6 \pm 0.2 $ for the hard stacking and rope-box tasks, respectively.

\begin{figure}
\centering
\includegraphics[width=\linewidth]{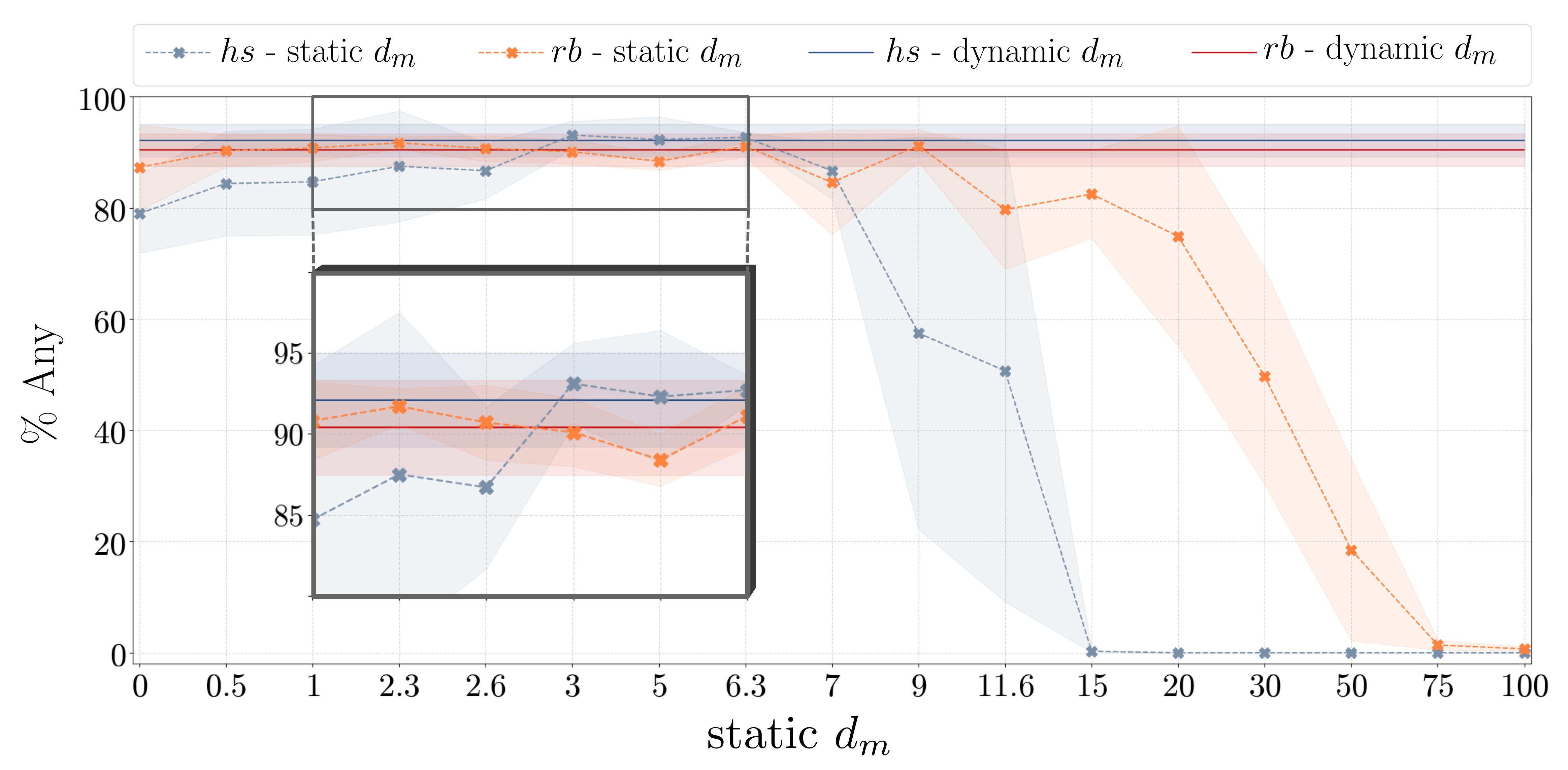}
 %\vspace{-15pt}
\caption{Comparison of LSR performance using the dynamic $d_m$ (solid lines) and static $d_m$ (cross markers with dashed lines) for the hard stacking (blue) and rope-box manipulation (orange) tasks. 
Non linear $x$-axis scale showing the values of $d_m$ is used for better visualization.}
\label{fig:sim:dm}
\end{figure}

 Figure~\ref{fig:sim:dm} shows the performance of the LSR using partial scoring on the hard stacking task (blue) and rope-box manipulation task (orange) obtained for the dynamic $d_m$ (solid lines), and a selected number of static $d_m$ parameters (cross markers with dashed lines) ranging from low ($d_m=0$) to high ($d_m=100$) values. Among the latter, we included the static $d_m=11.6$  and $d_m=6.3$  obtained using our previous approach in \cite{ouriros} on the stacking and the rope-box tasks, respectively.
We observed that: i) the choice of $d_m$ heavily influences the LSR performance, where same values of $d_m$ can lead to different behavior depending on the task (\eg, $d_m=11.6$), ii) the dynamic $d_m$ leads to nearly optimal performance regardless of the task compared to the grid searched static $d_m$. Note that even though there are static $d_m$ values where the performance is higher than in the dynamic case (\eg, $d_m=3$ with $93.1\%$ for stacking and $d_m=9$ with $91.2\%$ for the rope-box task), finding these values a priori without access to ground truth labels is hardly possible.

\textit{This approach not only eliminates the need for training the baseline VAEs as in \cite{ouriros}  but also reaches a value of $d_m$ that obtains a better  separation of covered regions $\mathcal{Z}_{sys}^i$ without compromising the optimization of the reconstruction and KL terms.}  
In fact, as discussed in Sec.~\ref{sec:mapping_module}, the reconstruction, KL and action terms in the loss function~\eqref{eq:vaeloss} have distinct influences on the latent space structure which can be in contrast to each other. The proposed dynamic increase of $d_m$ results in a lower $d_m$ value than in~\cite{ouriros}, which in turn yields small distances between the action pair states  while still being more beneficial than a simple static $d_m=0$. Such small distances in the action term are desirable as they do not contradict the KL term.
This can explain why the LSRs with higher values of $d_m$ reach worse performance compared to the dynamic one.  On the other hand, the quality of the obtained visual plans demonstrates that the resulting $d_m$ neither affects the reconstruction capabilities of the MM.

%-------------------------------------------------
%--------- gap analysis 
%-------------------------------------------------
\subsubsection{Separation of the states}
\label{sec:sec:stack:sep_states}
We investigated the effect of the action loss~\eqref{eq:acloss} on the structure of the latent space 
 by analyzing the separation of the latent points $z \in \mathcal{T}_z$ corresponding to different underlying states of the system. For simplicity, we report only results for the normal stacking task but we observed the same conclusions for the hard stacking and the rope-box manipulation tasks.
 Recall that images in $\mathcal{T}_I$ containing the same  state looked different because of the introduced positioning noise in the stacking tasks (and different lightning conditions in the case of $hs$ as well as the deformability of the rope in $rb$). 

Let $\bar z_s$ be the \emph{centroid} for state $s$ defined as the mean point of the training latent samples $\{z_{s, i}\}_i \subset  \mathcal{T}_z$ associated with the state $s$. Let $d_{intra}(z_{s, i}, \bar z_s)$ be the \emph{intra-state} distance defined as the distance between the latent sample~$i$  associated with the state~$s$, namely $z_{s, i}$,   and the respective centroid $\bar z_s$. Similarly, let $d_{inter}(\bar z_s, \bar z_p)$ denote the \emph{inter-state} distance between the centroids $\bar z_s$ and $\bar z_p$ of states $s$ and $p$, respectively.

\begin{figure}[htb!]
\begin{center}
\includegraphics[width=8.7cm]{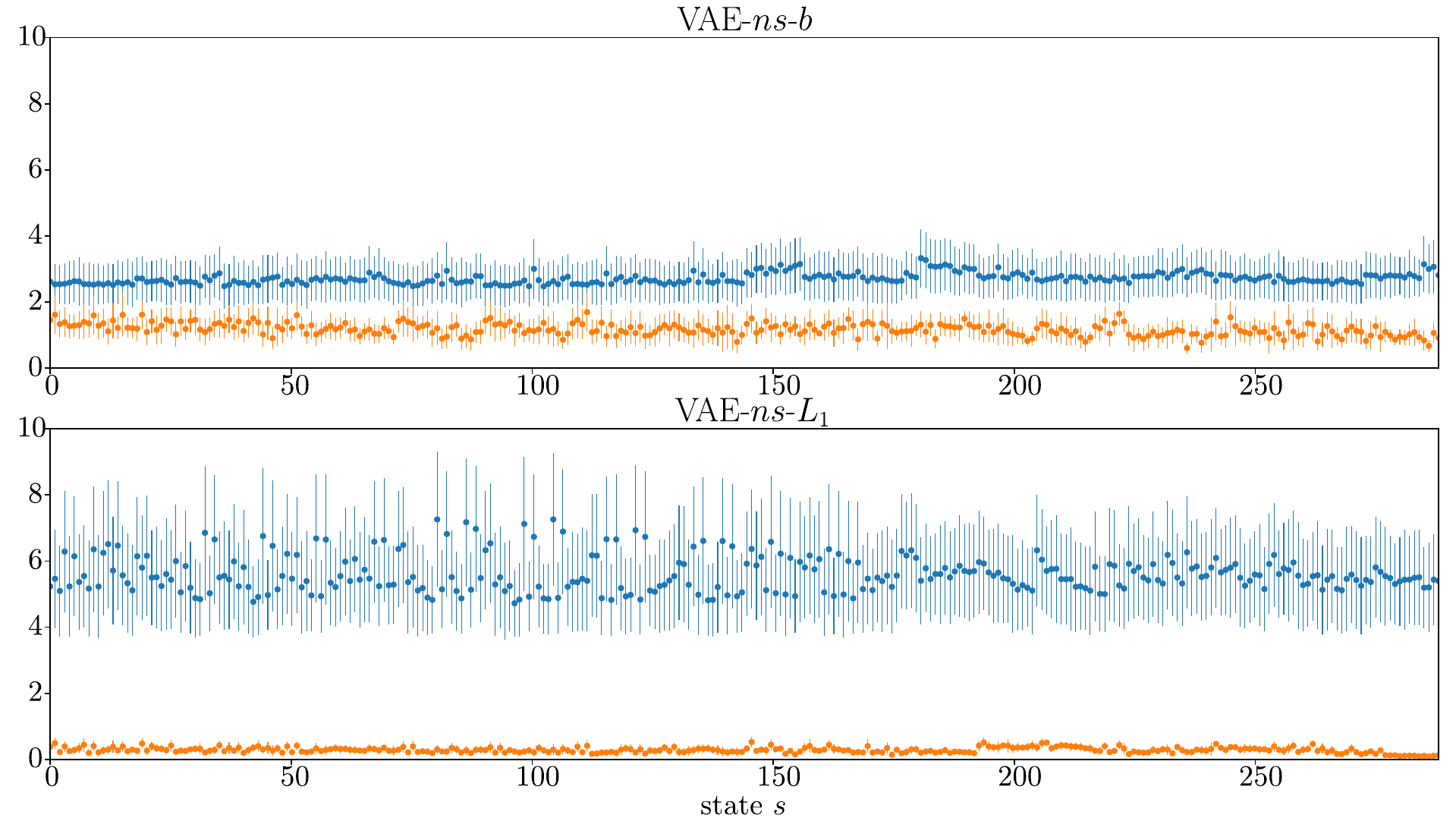}
\end{center}
\caption{Mean values (bold points) and standard deviations (thin lines) of inter- (blue) and intra- (orange) state distances for each state calculated using the baseline VAE (top) and the action $\vae{12}{ns}{L_1}$ model  (bottom) on normal stacking task.  }
\label{fig:sim1:dist-normal}
\end{figure}

Figure~\ref{fig:sim1:dist-normal} reports the mean values (bold points) and the standard deviations (thin lines) of the inter- (in blue) and intra-state (in orange) distances for each state $s\in\{1,...,288\}$ in the normal stacking task when using the baseline model $\vae{12}{ns}{b}$ (top) and the action model $\vae{12}{ns}{L_1}$ (bottom). 
In case of the baseline VAE, we  observed similar intra-state and inter-state distances. This implies that samples of different states
were encoded close together in the latent space which can raise ambiguities when planning.
On the contrary, when using $\vae{12}{ns}{L_1}$,  we 
observed that the inter- and intra-state distances approach the values $5$
and $0$, respectively. These values 
were imposed with the action term~\eqref{eq:acloss} as the minimum distance $d_m$  
reached $2.6$. 
Therefore, even when there  
 existed no direct link between two samples of different states, and thus the action term for the pair
was never activated, the VAE 
was able to encode them such that the desired distances in the latent space  
were respected.
Similar conclusions also hold for the hard stacking and the rope-box manipulation tasks, whose plots are omitted for the interest of space.

Finally, we analyzed the difference between the minimum inter-state distance and the maximum intra-state distance for each state. The higher the value the better separation of states in the latent space since samples of the same state are in this case closer to each other than samples of different states. 
When the latent states were obtained using the baseline $\vae{12}{ns}{b}$, we 
observed a non-negative distance for $0/288$ states
with an average value of $\approx -1.2$.  This implies that only weak separation  
 occurred in the latent space for samples of different states.  On the other hand, when calculated on points encoded with $\vae{12}{ns}{L_1}$, 
the difference became non-negative for $284/288$ states and its mean value 
 increased to $\approx 0.55$,  thus achieving almost perfect separation.
In the hard stacking task, we similarly found that $\vae{12}{hs}{b}$ reached an average difference of $-5.86$ (being non-negative for $0/288$ states), while the action model $\vae{12}{hs}{L_1}$ reduced the average difference to $-0.04$  (being non-negative for $121/288$ states). This result demonstrates the difference in the difficulty between the two versions of the box stacking task and  highlights the challenges of visual action planning on the harder stacking task where worse separation of states 
was achieved.  For the rope-box manipulation task we obtained, coherently with the box stacking results, an average difference of $-2.95$ (being non-negative for $37/157$ states) with the baseline model, which improved to $0.15$ with the action model  $\vae{12}{rb}{L_1}$ (being non-negative for $100/157$ states).   

In Appendix\ref{sec:sec:stack:ld_dim}, we performed an ablation study on the latent space dimension, justifying the choice $ld = 12$ in our simulations. %For each considered latent dimension, we then additionally analyzed the influence of the action loss on the structure of the latent space by calculating \textit{relative contrast}~\cite{aggarwal2001surprising, beyer1999nearest} in Appendix\ref{sec:sec:stack:rc}, which evaluates the significance of distances among points in the latent space. We observed higher values for action VAEs compared to the baseline VAEs, showing that the action loss leads to higher relative contrast in the latent space. 

\textit{
We conclude that the action term~\eqref{eq:acloss} and the dynamic $d_m$  contribute to a better structured latent space $\mathcal{Z}_{sys}$. }

\subsubsection{VAE compared to AE}
\label{sec:sec:stack:vae_vs_ae}
VAE framework is only one of the possible models for the MM. We justify this modeling choice by comparing it to the AE framework. Similarly as VAE, an AE model consists of an encoder and a decoder network which are jointly trained to minimize the the Mean Squared Error (MSE) between the original input and its decoded output. 
In contrast to VAEs, the two networks 
in AEs do not model a probability distribution.
Since the KL divergence in VAE  acts as a regularization term, we employed the stable weight-decay Adam optimizer from~\cite{xie2020stable} with default parameters to make the comparison more fair. We denote the model by AE-$b$.
Analogously to VAE, the original AE loss was augmented with the action loss~\eqref{eq:acloss} weighted by the parameter $\gamma$, which we denote by AE-$L_1$. Note that $L_1$ refers only to the metric in~\eqref{eq:acloss} and not in the MSE calculation.

We modelled the AE encoder and decoder networks using the same ResNet~\cite{ResNet} architecture as in case of VAEs. We set $ld = 12$, $\gamma = 1000$ and increased
the minimum distance $d_m$ dynamically every fifth epoch starting from $0$ using $\Delta d_m=1$, as described in Sec.~\ref{sec:mapping_module}. The LSR was built using the same  $\tau_{\min}=0$ and $\tau_{\max}=3$ (Algorithm~\ref{alg::dendrogram_cut}).

 Table~\ref{tab:ae_comparison_all} shows the LSR performance using partial scoring on all simulated tasks when MM was modelled as an AE (top two rows) and as a VAE (bottom row). 
Not only we observed a superior performance of VAE compared to the AE but once again the effectiveness of the action term~\eqref{eq:acloss}  on all the tasks  as it increased the average AE performance from  $0.1\%$ to  $36.3\%$ for $ns$, from  $0.1\%$ to  $33.6\%$ for $hs$, and  $0.1\%$ to  $9.8\%$ for $rb$.
\textit{
This comparison shows that the probabilistic modeling adopted by VAEs 
resulted in a latent space that is more adequate for visual action planning with respect to the considered AEs. }
As future work, we aim to investigate the benefits of more advanced models, such as Vector Quantised-VAE \cite{van2017neural}, which are out of the scope of this work.

\begin{table}[h!]
\resizebox{\linewidth}{!}{%
\centering
{ 
\begin{tabular}{|c|c|c|c|}
\hline
Model & $ns$ [\%]  & $hs$ [\%] & $rb$ [\%] \\ \hline
AE-$b$+LSR-$L_1$  &  $0.1 \pm 0.0$ & $0.0 \pm 0.0$ & $0.1 \pm 0.1$\\ \hline
AE-$L_1$+LSR-$L_1$ & $36.3 \pm 26.9$  &  $33.6 \pm 10.3$ & $9.8 \pm 5.4$ \\ \hline
VAE-$L_1$+LSR-$L_1$ &    \boldmath$100.0 \pm 0$ & \boldmath$92.1 \pm 2.9$ &  \boldmath{$90.4 \pm 2.9$} \\ \hline
\end{tabular}%
}
}
\caption{Comparison of the LSR performance using partial scoring when modelling MM with an AE (top two rows) and a VAE (bottom row) framework on all the simulated tasks. Best results in bold. }
\label{tab:ae_comparison_all}
\end{table}

%---------------------------------------------------------------
%----------------------------------------------------------------

\subsection{LSR Analysis}\label{sec:sim:lsr}

In this section, we analyze the LSR performance by answering the questions stated in point~\ref{en:sim:lsr}) of Sec.~\ref{sec:box-intro}. Firstly, we compared the LSR performance to the method in~\cite{savinov2018semiparametric} and one  inspired by~\cite{hafner2018learning}. Secondly, we investigated the influence of the action term~\eqref{eq:acloss} on the LSR performance.
Thirdly, we investigated the influence of the upper bound on the number of connected components $c_{\max}$ used in~\eqref{eq:brets_objective}. Next, we 
performed an extensive comparison of the LSR algorithm using different clustering algorithms in Phase $2$ of Algorithm~\ref{alg::fix_epsilon}.  Finally, we analyzed the covered regions determined by the LSR.

\subsubsection{LSR comparison}
\label{sec:sec:stack:lsr_performance}
 We compared the performance of the LSR on all simulated tasks with two benchmark methods introduced below.
In all the experiments, we considered the baseline models VAE$_{12}$-$b$ and the action VAE$_{12}$-$L_1$ trained with the action term~\eqref{eq:acloss}.
We compared our method with Semi-Parametric Topological Memory (SPTM) framework~\cite{savinov2018semiparametric}  discussed in Sec.~\ref{sec:related-work} and an MPC-based approach inspired by~\cite{hafner2018learning}. 

In SPTM, we connected action pairs (treated as one-step trajectories) and no-action pairs (considered temporarily close) in the latent memory graph. As in~\cite{savinov2018semiparametric}, we added $N_\text{sc}$
more \textit{shortcut} edges connecting the encodings that are considered closest by the retrieval network to the memory graph.
In the localization step, we used the median of $ k = 5$ nearest neighbours of the nodes in the memory graph as recommended in~\cite{savinov2018semiparametric}. To select the waypoint, 
we performed a grid search over $ s_{\text{reach}} \in \{0.75, 0.9, 0.95 \}$ and chose ${s_{\text{reach}} = 0.95}$.  We also performed 
a grid search over ${N_\text{sc} \in \{0, 2\cdot 10^2, 1\cdot 10^4, 1\cdot 10^5,  1\cdot 10^6, 1.5\cdot 10^6, 2\cdot 10^6 \}}$~and used the values $ N_\text{sc} =  1.0 \cdot 10^6,\, 1.5 \cdot 10^6,\, 2.0 \cdot 10^6 $ for $ns$, $hs$ and $rb$, respectively.
{We used high number of shortcuts compared to ${N_\text{sc} =  2 \cdot 10^2}$ in~\cite{savinov2018semiparametric} because we only had access to one-step trajectories instead of full roll-outs. Using low number of shortcuts resulted in a memory graph consisting of large amount of \emph{disconnected} components which impeded planning. For example, in hard stacking task using ${N_\text{sc} =  2 \cdot 10^2}$ yielded a graph with  $2243$  connected components 
which led to almost zero correct transitions over the $1000$ test paths. A higher number of shortcuts instead improved the connectivity of the graph and thus its planning capabilities.

The MPC-inspired baseline is composed of a learned transition model $f_t(\cdot)$ and a learned action validation model $f_a(\cdot)$, both taking the current latent state $z_1$ and the applied action $u$ as inputs. The transition model then predicts the next state $z_2=f_t(z_1,u)$, while the validation model $f_a(z_1,u)$  predicts whether the given action $u$ was allowed or not.

These models are used in a MPC-style approach, where first a search tree is constructed for a given start state $z_1$ by iterating over all allowed action using $f_a(z_1,u)$ with $u \in \mathcal{U}$ and predicting the consecutive states with the transition model $f_t(\cdot)$. The search is performed at each time step and until the search tree has reached a specified horizon $N$. Lastly, the path in the built tree leading to the state closest to the goal using $L_1$ distance is selected and the first action in the sequence is applied. 
This procedure is repeated until all proposed actions lead further from the goal.
In our case, the resulting state and action sequence is decoded into a visual action plan and evaluated in the same way as the LSR.

We implemented $f_t$ and $f_a$ 
as a three layer MLP-regressor and MLP-classifier, respectively, with $100$ hidden units. 
For a fair comparison, we trained $f_t$  and $f_a$ using training encodings $\mathcal{T}_z$ from the same MM that was used for building the LSR. 
As $\mathcal{T}_z$ only includes allowed actions, we augmented the training data for $f_a(\cdot)$ with an equal amount of negative examples by randomly sampling $u \in \mathcal{U}$. We used horizon $N=4$. 
The trained $f_t$ models achieved $R^2$ coefficient of determination~\cite{berk1990primer} of
$0.96$, $0.96$, and $0.88$ (highest $1$) for the normal, hard stacking and rope-box datasets, respectively.  
The $f_a(\cdot)$ model was evaluated on $1000$ novel states and by applying all possible actions on each state. It achieved an accuracy score of $88.5 \pm 1.8$, $97.3 \pm 0.2$, and $87.4 \pm 0.8$ for the normal, hard stacking and rope-box datasets, respectively. 
Note that the normal and hard stacking tasks has exactly $48$ unique actions with $\approx 9.4\%$ of them being allowed on average. The rope-box task on the other hand has $25$ unique actions with an average of $\approx 17.1\%$ being allowed per state. 

Table~\ref{tab:sim_lsr_eval} shows the result of our method (VAE-$L_1$ + LSR-$L_1$), 
the SPTM framework and the MPC-based approach (VAE-$L_1$ + MPC) evaluated on the full scoring on the normal box stacking (top), hard box stacking (middle), and rope-box manipulation task (bottom).}
 We observed that the proposed approach (VAE-$L_1$ + LSR-$L_1$) significantly outperformed the considered benchmark methods. This can be explained by the fact that SPTM- and MPC-based methods are more suited for tasks where the provided data consists of rolled out \textit{trajectories} in which small state changes are recorded in consecutive states, which is also a potential shortcoming of~\cite{liu2020hallucinative}.

In contrast, as discussed in Sec.~\ref{sec:limit},    our method is best applicable when actions lead to distinguishable different observations. This allows to consider only  pairs of observations as input dataset instead of requiring entire trajectories. Moreover, a core difference between our approach and SPTM is that we do not assume that each observation maps into a unique underlying state, but rather, as described in Sec.~\ref{sec:overview}, we structure and cluster observations in such a way that observations associated with the same underlying state are grouped together. We reiterate that  this approach is best suited for tasks with finite and distinguishable states, 
which differ from continuous RL setting used by SPTM.

\subsubsection{Influence of the action term}
\label{sec:sec:stack:lsr_action_loss}
We investigated how the LSR performance is affected by the action term~\eqref{eq:acloss} by comparing it to the variant where MM was trained without it (VAE-$b$ + LSR-$L_1$). The results on the full scoring for all the tasks are shown in Table~\ref{tab:sim_lsr_eval}. We observed deteriorated LSR performance when using baselines VAE$_{12}$-$b$ compared to the action VAEs regardless the task.  This  indicates that  VAEs-$b$  were not  able  to  separate  states  in $\mathcal{Z}_{sys}$. We again conclude that the action term~\eqref{eq:acloss} needs to be included in the VAE loss function~\eqref{eq:loss} in order to obtain distinct  covered regions~$\mathcal{Z}_{sys}^i$. 
In addition, the results  underpin the different level of difficulty of the  tasks as indicated by the drop in the LSR performance on $hs$ and $rb$ compared to $ns$ using the action VAE-$L_1$.
 
\textit{In summary, this simulation campaign  demonstrates the effectiveness of the LSR  on all the considered simulated tasks involving both rigid and deformable objects compared to existing solutions, as well as  supports the 
integration of the action term in the VAE loss function. }

\begin{table}[h!]
\resizebox{\linewidth}{!}{%
{ 
\begin{tabular}{|c|c|l|l|l|l|l|}
\hline
Task & Model  & \% All       & \% Any & \% Trans.  \\ \hline
\multirow{4}{*}{$ns$}  &  VAE-$L_1$ + MPC & $2.3 \pm 0.3$ & $2.3 \pm 0.3$ & $69.3 \pm 1.0$ \\ \cline{2-5}
 & SPTM~\cite{savinov2018semiparametric} &  $0.2 \pm 0.1$ & $ 0.5 \pm 0.3 $ & $51.9 \pm 1.4 $ \\ \cline{2-5}
 & VAE-$b$+ LSR-$L_1$ & $2.5 \pm 0.5$ & $4.1 \pm 1.0$ & $59.7 \pm 4.9$ \\ \cline{2-5}
 &  VAE-$L_1$+ LSR-$L_1$ &   {\boldmath$100.0 \pm 0$} & {\boldmath$100.0 \pm 0$} & {\boldmath$100.0 \pm 0$}  \\ \hline \hline
\multirow{4}{*}{$hs$} &  VAE-$L_1$ + MPC &  $2.1 \pm 0.4$ & $2.1 \pm 0.4$ & $76.8 \pm 0.3$
\\ \cline{2-5}
 & SPTM~\cite{savinov2018semiparametric} & $0.0 \pm 0.0 $ & $0.0 \pm 0.0$ & $23.6 \pm 0.7$  \\ \cline{2-5}
 & VAE-$b$+ LSR-$L_1$ & $0.2 \pm 0.1$ & $0.2 \pm 0.1$ & $38.0 \pm 2.0$ \\ \cline{2-5}
 & VAE-$L_1$+ LSR-$L_1$ &  \boldmath$90.9 \pm 3.5$ &  \boldmath$92.1 \pm 2.9$ &  \boldmath$95.8 \pm 1.3$  \\ \hline \hline
 \multirow{4}{*}{$rb$} &  VAE-$L_1$ + MPC & $6.2 \pm 0.5$ & $6.2 \pm 0.5$ & $73.8 \pm 0.8$ \\ \cline{2-5}
 & SPTM~\cite{savinov2018semiparametric} &  $0.0 \pm 0.0$ & $0.4 \pm 0.3$ & $25.2 \pm 9.7$ \\ \cline{2-5}
 &  VAE-$b$+ LSR-$L_1$ & $0.2 \pm 0.1$ & $0.2 \pm 0.1$ & $0.2 \pm 0.1$ \\ \cline{2-5}
 & VAE-$L_1$+ LSR-$L_1$ &  \boldmath$89.7 \pm 3.7$ & \boldmath$90.4 \pm 2.9$ & \boldmath$96.2 \pm 1.5$ \\ \hline
\end{tabular}%
}
%\end{adjustbox}
}
\caption{Planning performance using full scoring for the normal (top) and hard (middle) box stacking tasks and rope-box manipulation task (bottom) using MPC and SPTM~\cite{savinov2018semiparametric} methods, baseline VAE-$b$ and action VAE-$L_1$. Best results in bold.  }
\label{tab:sim_lsr_eval}
\end{table}

\subsubsection{Influence of the maximum number of connected components}
\label{sec:sec:stack:lsr_cmax}
The optimization method described in Sec.~\ref{ssec:LSR_optim} requires setting an upper bound on the number of graph-connected components $c_{\max}$ of the LSR. Table \ref{tab:stacking_eval_cmax} shows how different choices of upper bounds influence the LSR performance on  all simulated tasks.

\begin{table}[h!]
%\resizebox{\textwidth}{!}{%
\centering
{ 
\begin{tabular}{|c|c|c|c|}
\hline
$ c_{max}$ & $ns$ [\%] & $hs$ [\%] & $rb$ [\%] \\ \hline
$1$ & {\boldmath $100.0 \pm 0.0$ }& $65.3 \pm 24.6$ & $4.5 \pm 5.6$  \\ \hline
$5$ & $99.5 \pm 0.4$ & $88.6 \pm 5.4$ & $55.8 \pm 28.8$ \\ \hline
$10$ & $99.0 \pm 0.3$ & $91.5 \pm 3.8$ &  $80.4 \pm 10.6$\\ \hline
$20$ & $97.5 \pm 0.5$ & {\boldmath$92.1 \pm 2.9$ } & \boldmath$90.4 \pm 2.9$\\ \hline
$50$ & $91.3 \pm 1.1$ &  $88.2 \pm 2.0$ & $89.4 \pm 1.9$\\ \hline
$100$ & $80.0 \pm 1.4$ & $77.9 \pm 2.1$ & $76.0 \pm 2.8$\\ \hline
\end{tabular}%
%}
}
\caption{LSR performance on all simulated tasks for different $c_{max}$ values. Best results in bold. }
\label{tab:stacking_eval_cmax}
\end{table}

We 
observed that the results are rather robust with respect to the $c_{\max}$ value.  For  all tasks, the performance 
dropped for a very high $c_{\max}$, such as $c_{\max} = 100$, while in the hard stacking task  and especially in  the rope-box manipulation task, we additionally 
observed a drop for a very low $c_{\max}$, such as $c_{\max} = 1$.
This behavior can be explained by the fact that the lower the $c_{\max}$ the more the system is sensitive to outliers, while the higher the $c_{\max}$ the greater the possibility that the graph is disconnected which potentially compromises its planning capabilities. For example, in the hard stacking task, outliers arise from different lightning conditions, while in the rope-box manipulation task they arise from the deformability of the rope. In contrast, no outliers exist in the normal stacking task which is why a single connected component is sufficient for the LSR to perform perfectly. For all further evaluation,  we set $c_{max}=1$ for $ns$ and $c_{max}=20$ for $hs$ and $rb$. 

 \textit{This result demonstrates the robustness of the approach with respect to $c_{\max}$ as well as justifies the choices of the $c_{max}$ values in the rest of simulations.}

\subsubsection{Comparing different clustering methods for Phase 2}
\label{sec:sec:stack:lsr_phase2}

 We showcase the effect of the outer optimization loop described in Algorithm~\ref{alg::dendrogram_cut} on several different clustering methods used in Phase 2 in Algorithm~\ref{alg::fix_epsilon} on 
the hard stacking task. We considered \textit{Epsilon clustering} used in our earlier work~\cite{ouriros}, 
\textit{Mean-shift}~\cite{cheng1995mean}, OPTICS~\cite{ankerst1999optics}, Linkage (single, complete and average)~\cite{day1984efficient} and HDBSCAN~\cite{mcinnes2017hdbscan} algorithms. We provide a summary of the considered algorithms
in Appendix\ref{sec:app:clus-ae}. 
The performance of the 
considered clustering methods (except for HDBSCAN) depends on a single input scalar parameter that is hard to tune.  However, as described in Sec.~\ref{ssec:LSR_optim},  we are able to optimize it by maximizing the objective in~\eqref{eq:brets_objective}.

Table~\ref{tab:hardstacking_clusster_eval}  reports the  LSR performance with different clustering algorithms when performing grid search to determine their input scalar parameters (left) and when using our automatic optimization (right). Partial scoring 
using $\vae{12}{hs}{L_1}$ is shown. 
Note that the grid search 
was only possible in this problem setting as the ground truth can be retrieved from the trained classifiers but it is not generally applicable.
Firstly, the results show that average-linkage, used for our LSR in Sec.~\ref{ssec:LSR_building}, together with our automatic input parameter optimization 
outperformed the other alternatives. 
The results of the grid search show that the automatic criteria for identifying different cluster densities, adopted by OPTICS and HBDSCAN, 
 did not effectively retrieve the underlying covered regions. Meanshift % 
 performed better but its approximation of spherical clusters 
did not lead to the optimal solution. Similar performance to Meanshift  
was obtained with single- and complete-linkage algorithms showing that the respective distance functions are not either suited for identifying covered regions. The same applies for the epsilon clustering.

Concerning the optimization results, they highlight the effectiveness of the optimization procedure in Algorithm~\ref{alg::dendrogram_cut} as they are comparable to the ones obtained with the grid search for all clustering methods.
Note that grid search  
 led to a slightly lower performance than the optimization for meanshift, complete-linkage and average-linkage.
In these cases, the grid 
was not fine enough which points out the difficulty of tuning the respective parameters.

\textit{This investigation demonstrates the effectiveness of our proposed optimization loop and shows that the average-linkage clustering algorithm  
 led to the best LSR performance  among considered alternatives for the hard box stacking task.}

\begin{table}[h!]
\centering
%\resizebox{\linewidth}{!}{%
\begin{tabular}{|l|c|c|c|}
\hline
Clust. method & Grid Search [$\%$] & Optimization [$\%$] \\ \hline
Epsilon \cite{ouriros} &  $83.5 \pm 4.8$ & $65.8 \pm 12.2$
  \\ \hline
Meanshift  & $78.2 \pm 3.3$ & $80.2 \pm 5.9$ 
  \\ \hline
OPTICS &   $44.3 \pm 8.7$  & $40.8 \pm 6.1$ 
  \\ \hline
HDBSCAN &  $16.1 \pm 5.7$ & - \\ \hline 
Single-linkage  &  $79.3 \pm 8.8$ & $65.8 \pm 12.2$  \\ \hline
Complete-linkage &   $79.1 \pm 6.4$ & $81.4 \pm 4.8$  \\ \hline
Average-linkage &  $91.1 \pm 2.5$ &  {\boldmath $92.1 \pm 2.9$}  \\ \hline
\end{tabular}%
%}
\caption{Comparison of the LSR performance for different clustering algorithms for the hard box stacking task.  Partial scoring is reported when applying grid search (left column) and when using the optimization in Algorithm~\ref{alg::dendrogram_cut} (right column). Best results in bold. }
\label{tab:hardstacking_clusster_eval}
\end{table}

\subsubsection{Covered regions using LSR}
\label{sec:sec:stack:lsr_cover}
To show that the LSR captures the structure of the system, 
we checked if observations corresponding to true underlying states of the system, that have not been seen during training, are properly recognized as covered. Then, we  checked if observations from  the datasets of the remaining simulated tasks as well as from the   3D Shapes dataset~\cite{3dshapes18} are marked as uncovered since they correspond to out-of-distribution observations. 
The covered regions $\mathcal{Z}^i_{sys}$ were computed using the epsilon approximation in \eqref{eq:cluster_eps}.

Table~\ref{tab:stacking_valid_5seeds}
reports the results of the classification of covered states obtained by the models trained on normal (first row) and hard (second row) box stacking tasks and rope-box manipulation task (third row).
 Holdout datasets for each simulated task were used. 
For the normal stacking task,  
 results show that the LSR almost perfectly recognized all the covered states ($ns$ column)  with the average recognition equal to $99.5\%$, while it properly recognized on average $4694/5000$ samples ($93.9\%$ - $hs$ column) hard version.  An almost perfect  average recognition was also obtained on the rope-box manipulation task ($99.6\%$ - $rb$ column). 
For out-of-distribution observations, 
the lower the percentage the better the classification.   Table~\ref{tab:stacking_valid_5seeds} shows that the models trained on $ns$ (first row, columns $hs$, $rb$, $3$D Shapes) and $hs$ (second row, columns $ns$, $rb$, $3$D Shapes) were able to perfectly identify all \textit{non}-covered states,  
while worse performance was observed for the rope-box models which misclassified $\approx 10\%$ of the uncovered datasets (third row, columns $ns$, $hs$, $3$D Shapes). This decrease in performance could be explained by the fact that capturing the state of a deformable object is much more challenging than rigid objects.

\textit{We conclude that LSR provides a good approximation of
the global structure of the system as it correctly  classified most of the observations representing possible system states as covered, and out-of-distribution observations as not covered. }

\begin{table}[h!]
\centering
\resizebox{\linewidth}{!}{%
{ 
\begin{tabular}{|c|c|c|c|c|c|}
\hline
  & $ns$ [$\%$] & $hs$ [$\%$] & $rb$ [$\%$] &  3D Sh. [$\%$] \\ \hline
{$ns$}  & $99.47 \pm 0.27$ &  $0.0 \pm 0.0$ & $0.0 \pm 0.0$ &  $0.0 \pm 0.0$ \\ \hline
{$hs$} & $0.0 \pm 0.0$ & $ 93.71 \pm 0.61$  & $0.0 \pm 0.0$ & $0.0 \pm 0.0$  \\ \hline
{$rb$} & $ 9.48 \pm 7.45$ & $13.5 \pm 8.57$ & $ 99.6 \pm 0.1$ & $9.72 \pm 8.38$
  \\ \hline
\end{tabular}%
}
}
\caption{Classification of covered states for the normal (first row) and hard (second row) box  stacking models and rope-box models (third row) when using as inputs novel images from the tasks ($ns$, $hs$ and $rb$ columns) and the  3D Shapes  ($3$D Sh. column) datasets. }
\label{tab:stacking_valid_5seeds}
\end{table}

%-------------------------------------------------
%---------APN
%-------------------------------------------------

\subsection{APM Analysis}
\label{sec:fold:apm}
We evaluated the accuracy of action predictions obtained by APN-$L_1$ on an unseen holdout set consisting of  $1611$, $1590$ and  $948$ action pairs for the normal stacking,  hard stacking and rope-box manipulation tasks, respectively. As the predicted actions can be binary classified as either true or false, we calculated the percentage of the correct proposals for picking and releasing, as well as the percentage of pairs where both pick and release proposals were correct. For rope-box task, we additionally calculated the percentage of the correct proposal for either rope or box action. We evaluated all the models on $5$ different random seeds. 
For both stacking versions, all the models performed with accuracy $99\%$ or higher, while rope-box models achieved $\approx 96\%$. This is because the box stacking task results in an $18$-class classification problem for action prediction which is simple enough to be learned from any of the VAEs, while the classification task in the rope-box is slightly more challenging due to the required extra prediction whether to move a rope or a box.

%-------------------------------------------------------------------------------
%-------------------------------------------------------------------------------
%-------------------------------------------------------------------------------
%-------------------------------------------------------------------------------

\section{Folding Experiments}\label{sec:exp}
In this section, we validate the proposed approach on a real world experiment involving  manipulation of deformable objects, namely folding a T-shirt. As opposed to the  simulated tasks, the true underlying states 
were in this case unknown and it % 
was therefore not possible to define an automatic verification of the correctness of a given visual action plan.

The folding task setup is depicted in Fig.~\ref{fig:folding:super_results_folding} (middle). We used a Baxter robot equipped with a Primesense RGB-D camera 
mounted on its torso to fold a T-shirt in different ways.
The execution videos of all the performed experiments and respective visual action plans can be found on the project  website. A summary of the experiments can also be found in the accompanying video.  
For this task, we collected a dataset $\mathcal{T}_I$ containing $1283$ training tuples. Each tuple consists of two images of size $256\times256\times 3$, and action specific information $u$ defined as $u=(p,r,h)$ where $p=(p_r, p_c)$ are the picking coordinates,  $r=(r_r, r_c)$  the releasing coordinates and  $h$ picking height. 
 An example of an action and a no-action pair is shown in Fig.~\ref{fig::act-no-ac-fold}.
The values $p_r, p_c,r_r,r_c \in \{0,\dots, 255\}$ correspond to image coordinates, while $h \in \{0,1\}$ is either the height of the table or a 
value measured from the RGB-D camera to  pick up only the top layer of the shirt.
Note that  the separation of stacked clothing layers is a challenging task and active research area on its own~\cite{qian2020cloth}
 and leads to decreased performance when it is necessary to perform it, as shown in Sec.~\ref{sec:sec:fold:m_layers}. 
%
%The dataset $\mathcal{T}_I$ was collected by manually selecting {\red relevant}  pick and release points on images showing a given T-shirt configuration {\blue demoing different folds}, and recording the corresponding action and following configuration.
The dataset $\mathcal{T}_I$ was collected by providing task demonstrations by human operators, i.e., by manually selecting pick and release points on images showing a given T-shirt configuration, and recording the corresponding action and following configuration.
No-action pairs, representing $\approx37\%$ of training tuples in $\mathcal{T}_I$, were generated by slightly perturbing the cloth appearance.

\subsection{Experiment Objectives and Implementation Details} \label{sec:folding:details}
The experiments on the real robot %
were designed to answer the following questions:

\begin{enumerate}
    \item \textbf{MM} Does the action loss term~\eqref{eq:acloss} improve the structure of the latent space for the folding task?
    
    \item \textbf{LSR} How good is the approximation of the covered regions  provided by the LSR for a real world dataset? 
    
    \item \textbf{APM} How does the APN perform in comparison to alternative implementations of the APM?
    
    \item \textbf{System} How does the real system perform and how does it compare to our earlier work~\cite{ouriros}? 
    What is the performance on a folding that involves picking the top layer of the shirt? 

\end{enumerate}

 Following the notations introduced in Sec.~\ref{sec:box-intro}, we denote by $\vae{ld}{f}{d}$ a VAE with $ld$-dimensional latent space, where $f$ stands for the folding task and $d$ indicates whether or not the model was trained with the action loss~\eqref{eq:acloss}. We use $d = b$ for the \textit{baseline} VAEs which were trained with the original training objective~\eqref{eq:vaeloss}.  We use $d = L_p$ for the \textit{action} VAEs trained with the objective~\eqref{eq:loss} containing the action term~\eqref{eq:acloss} using metric $L_p$ for $p \in \{1, 2, \infty\}$. 
We  modelled VAEs with the same ResNet architecture and same hyperparameters $\beta$, $\gamma$ and $d_m$ as in the box stacking task introduced in Sec.~\ref{sec:box} but increased the latent space dimension to $ld = 16$. We refer the reader to the code repository\footrefc{fn:git} for implementation details.

For the LSR,  we denote by $\lsr{p}$  
a graph obtained by using metric $L_p$ in Algorithm~\ref{alg::fix_epsilon}. We set the upper bound $c_{\max}$ 
in~\eqref{eq:brets_objective} to $5$, and the search interval boundaries $\tau_{\min}$ and $\tau_{\max}$ in Algorithm~\ref{alg::dendrogram_cut} to $0$ and $3.5$, respectively. 

The performance of the APMs and the evaluation of the system  
was based on the $\vae{16}{f}{L_1}$ realization of the MM.
We therefore performed the experiments using $\apn{16}{f}{L_1}$ which was trained on latent action pairs $\overline{\mathcal{T}_z}$ extracted by the latent mapping $\xi$ of $\vae{16}{f}{L_1}$. 
We trained $5$ models for $500$ epochs using different random seeds as in case of VAEs, and used $15\%$ of the training dataset as a validation split to extract the best performing model for the evaluation.

\begin{figure}[htb!]
\begin{center}
\includegraphics[width=7.5cm]{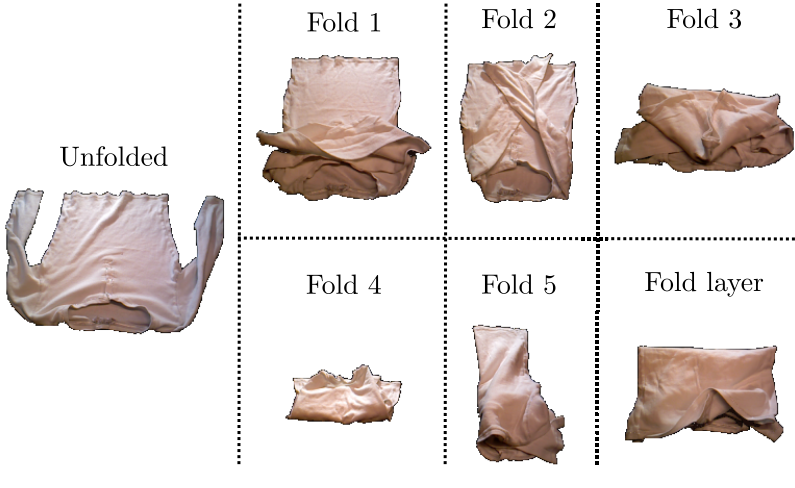}
\end{center}
\caption{Start state (right) followed by $5$ different goal configurations for the folding task~\cite{ouriros}. The lower right configuration requires to pick a layer on top of the T-shirt. }
\label{fig:folding_s_g}
\end{figure}

We compared the performance of our system S-OUR consisting of $\vae{16}{f}{L_1}$, $\lsr{1}$ and $\apn{16}{f}{L_1}$ with the systems S-${L_1}$, S-${L_2}$ and S-$L_{\infty}$ introduced in~\cite{ouriros},  using metrics $L_1$, $L_2$ and $L_{\infty}$, respectively, on the same folding tasks.  The major novelties of S-OUR with respect to the systems in~\cite{ouriros} are reported in Sec.~\ref{sec:intro}.  
The start configuration was the fully unfolded shirt shown in Fig.~\ref{fig:folding_s_g} on the left, while the $5$ goal configurations are shown on the right. The latter are of increasing complexity requiring a minimum of $2$, $2$, $3$, $3$, and $4$ folding steps for folds $1$-$5$, respectively.

\begin{figure*}[!h]
\vspace{7pt}
\centering
\includegraphics[width=0.9\textwidth]{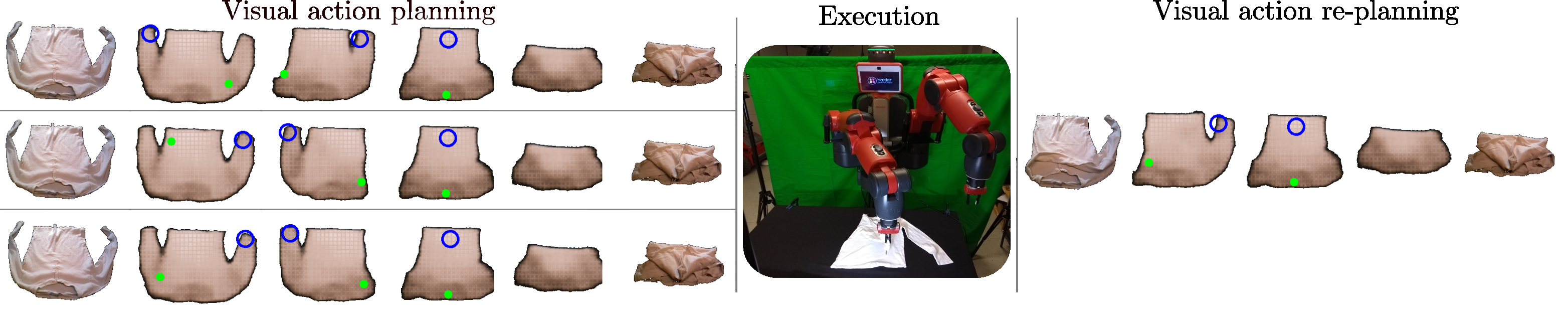}
\caption{Execution of the folding task with re-planning. On the left, a set of initial visual action plans reaching the goal state is proposed. After the first execution, only one viable visual action plan remains. }
\label{fig:folding:super_results_folding}
\end{figure*}

Each fold was repeated $5$ times and scored in the same way as in~\cite{ouriros}. In particular, we scored the \textit{system performance} where a folding was considered successful if the system was able to fold the T-shirt into the desired goal configuration. As the state space of the T-shirt is high-dimensional, there exists no objective measure that would evaluate the success of the fold automatically.
Therefore, the evaluation of the full folding procedure was  manually done by a human (one of the authors) but all execution videos of all folds and repetitions can be found on the project website. We additionally evaluated the percentage of successful \textit{transitions of the system}. A transition was considered successful if the respective folding step was executed correctly. Lastly, we evaluated the quality of the generated visual plans $P_I$ and the generated action plans $P_u$. We considered a visual (action) plan successful if all the intermediate states (actions) were correct. 
Even for a correctly generated visual action plan, the open loop execution is not robust enough for a real robot system. We therefore added a re-planning step after each action completion as shown in Fig.~\ref{fig:folding:super_results_folding}. This accounts, as instance,  for potential execution uncertainties, 
inaccuracies in grasping or in the positioning phases of pick-and-place operations which  
led to observations different from the ones planned in $P_I$. Note that after each action execution, the current observation of the cloth was considered as a new start observation, and a new visual action plan was produced until the goal observation is reached or the task is terminated.  Such re-planning setup was used for all folding experiments. 
As the goal configuration does not allude to how the sleeves should be folded, the LSR suggests multiple latent plans. 
A subset of the corresponding visual action plans is shown on the left of Fig.~\ref{fig:folding:super_results_folding}. If multiple plans were generated, a human operator selected one to execute. After the first execution, the ambiguity arising from the sleeve folding was removed and the re-planning generated a single plan, shown in the right.

To deal with the sparse nature of the collected dataset, if no path was found from the start to the goal node, the planning was repeated using the closest nodes to the current start and/or goal nodes in the latent space. This procedure was repeated until a path 
was found.

%-------------------------------------------------
%---------rc shirt
%-------------------------------------------------

\subsection{MM Analysis}

% Analogously to the box stacking task in Sec.~\ref{sec:sim:mm}, 
We answered question 1) %by  analyzing the relative contrast (reported in Appendix\ref{sec:sec:stack:rc}) as well as 
by evaluating the separation of action and no-action pairs during the training.

\subsubsection{Influence of dynamic $\boldsymbol{d_m}$}
\label{sec:sec:fold:dm}
\begin{figure}[htb!]
\begin{center}
\includegraphics[width=\linewidth]{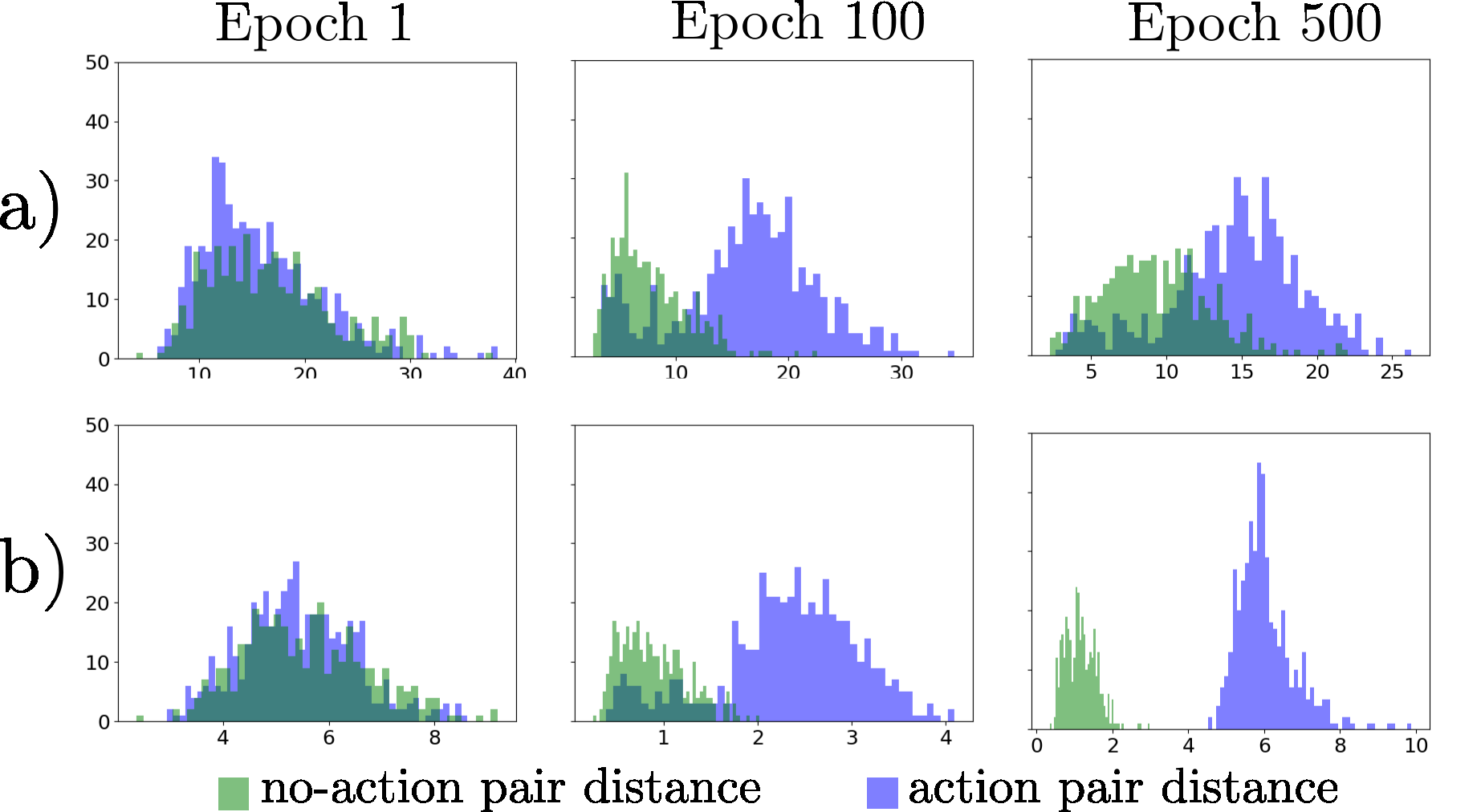}
\end{center}
\caption{Histograms of action (in blue) and no-action  (in green) pair distances at different training epochs ($1$, $100$ and  $500$ from the left, respectively) for the folding task. Results obtained with  baseline (top, a)) 
and action  (bottom, b)) models are shown.
}
\label{fig::d_m_histo_shirt}
\end{figure}

We investigated the influence of the dynamic increase of $d_m$ in the action term~\eqref{eq:acloss} on the structure of the latent space.
Figure~\ref{fig::d_m_histo_shirt} shows the histogram of action (in blue) and no-action (in green) pair distances calculated at different epochs during training using $\vae{16}{f}{b}$ (top row) and $\vae{16}{f}{L_1}$ (bottom row). 
The figure shows that the separation was complete in case of action VAEs but was not achieved with the baseline VAEs. To precisely quantify the amount of overlap between action and no-action pairs, we calculated the difference between the minimum action-pair distance and maximum no-action pair distance on the training dataset,  that is reported in the following. A positive  difference value implies that action pairs were successfully separated from the no-action pairs.
For $\vae{16}{f}{b}$ (top row), the difference evaluated to  $-31.8$,  $-19.2$, and $-19.4$ for epoch $1$, $100$, and $500$, respectively, while it was improved to $-6.3$, $-1.6$, and $1.5$ in case of the action $\vae{16}{f}{L_1}$ (bottom row). 
\textit{This shows that the dynamic selection of $d_m$ successfully 
 separated the actions and no-action pairs also for the folding task.}

%-------------------------------------------------
%---------LSR covered states shirt
%-------------------------------------------------

\begin{table*}[!h]
%\resizebox{\linewidth}{!}{%
\centering
\begin{tabular}{|c|c|c|c|c|c|c|}
\hline
Method & X Pick & Y Pick & X Release & Y Release & Height & Total \\ \hline
e-APN~\cite{ouriros} & $ 144.1 \pm 52.2 $  & $ 52.8 \pm 18.3 $  &  $ 317.2 \pm 143.3 $  &  $ 159.9 \pm 17.4 $ &  {\boldmath$0.0 \pm 0.0$}  &  $ 674.0 \pm  147.6 $\\ \hline
C-APN & $498.0 \pm 63.8$  & $ 47.4 \pm 7.7 $  &  $ 818.8 \pm 121.9 $  &  $ 226.5 \pm 92.5 $ &  {\boldmath$0.0 \pm 0.0$}  &  $ 1590.8 \pm 155.0 $\\ \hline
R-APN & $ 697.2 \pm 345.1 $  & $ 246.2 \pm 174.9 $  &  $ 792.4 \pm 388.8 $  &  $ 268.9 \pm 157.0$ & {\boldmath$0.0 \pm 0.0$}  &  $ 2004.6 \pm 908.2$\\ \hline
AAB & $113.0$ & {\boldmath$22.4$} & {\boldmath$201.4$} & $194.7$ & {\boldmath$0.0$} & $531.5$ \\ \hline
APN (Ours) & {\boldmath$82.6 \pm 22.9$}  & $29.3 \pm 2.2$  &  $270.6 \pm 158.2$  &  {\boldmath$71.8 \pm 15.0$} &  {\boldmath$0.0 \pm 0.0$}  &  {\boldmath$454.3 \pm 153.8$}\\ \hline
\end{tabular}%
%}
\caption{Comparison of MSE achieved with different realizations of the Action Proposal Modules. Best results in bold.}
\label{tab:apn_basline_comp}
\end{table*}

\subsection{LSR Analysis}\label{sec:exp:lsr}

Similarly to the simulated tasks, we exploited the LSR to investigate the covered regions of the latent space $\mathcal{Z}$, thus answering question 2) listed in Sec.~\ref{sec:folding:details}.  Note that in Sec.~\ref{sec:exp:real_robot}, the LSR was also employed to perform the folding task with the real robotic system.

\subsubsection{Covered regions using LSR}
\label{sec:sec:fold:lsr_cover}
We used $\vae{16}{f}{L_1}$ model and reproduced the experiment from Sec.~\ref{sec:sec:stack:lsr_cover}, where we measured the accuracy of various novel observations being recognized as covered. We 
 inputted $224$ novel observations that correspond to possible states of the system not used during training, as well as  $5000$ out-of-distribution samples from each of the three datasets of the simulated tasks and the standard 3D Shapes dataset.    We  
observed that the LSR achieved good recognition performance even in the folding task. More precisely, on average $213/224$ samples representing true states of the system were correctly recognized as covered, resulting in $95 \pm 2.4 \%$ accuracy averaged over the $5$ different random seeds.  For the four out-of-distribution datasets, all samples were correctly recognized as not covered.

\textit{This analysis illustrates
the effectiveness of the LSR in capturing the covered regions of the latent space.  
}

%-------------------------------------------------
%---------APN shirt
%------------------------------------------------

\subsection{APM Comparison}
\label{sec:exe:apm}
In this section we validate the choice of the APM by comparing it to several possible alternatives.

The Action Proposal Network, described in Sec.~\ref{sec:apm},  
was built upon the one introduced in~\cite{ouriros} to which we added dropout regularization layers. The APN receives as inputs latent action pairs contained in a latent plan found by the LSR, and outputs the predicted action specifics.  We refer to the \textit{earlier} version in~\cite{ouriros} as \textit{e-APN} and to the current version $\apn{16}{f}{L_1}$ as \textit{APN}.  We compared the performance of APN to e-APN as well as several alternatives introduced below.

\noindent \textbf{Action Averaging Baseline (AAB)} Firstly, we investigated whether the action predictions can be retrieved directly from the LSR instead of a separate module. The basic idea is to use the latent action pairs in the training dataset to calculate the average action specifics associated with each edge in the LSR. 
Let $\mathcal{E}_{sys}^{ij} = \{(z_1, z_2) \in \mathcal{E} | z_1 \in \mathcal{Z}_{sys}^i, z_2 \in \mathcal{Z}_{sys}^j \}$
be the set of edges from the reference graph $\mathcal{E}$ connecting covered regions $\mathcal{Z}_{sys}^i$ and $\mathcal{Z}_{sys}^j$ (Algorithm~\ref{alg::fix_epsilon}). We parameterized each edge $e_{\LSR}^{ij} = (z_{\LSR}^i, z_{\LSR}^j) \in \mathcal{E}_{\LSR}$  with the action $u_{\LSR}^{ij}$ obtained by averaging actions corresponding to the edges in~$\mathcal{E}_{sys}^{ij}$ 
\begin{equation}\label{eq:act-baseline}
    u_{\LSR}^{ij} = \frac{1}{|\mathcal{E}_{sys}^{ij}|}\sum_{(z_1, z_2) \in \mathcal{E}_{sys}^{ij}} u^{z_1z_2}
\end{equation}
where $u^{z_1z_2}$ is the action specification associated with the action pair $(z_1,z_2)$ in the training dataset $\mathcal{T}_z$. The parametrization~\eqref{eq:act-baseline} yields the action plan associated with a path $P_z$. 

Secondly, we investigated how the use of the latent encodings as inputs to the APM influences the LSR performance. We compared APN-d with two distinct versions of APMs that use images as inputs.\\
\noindent \textbf{C-APN} is a neural network that uses a convolutional encoder followed by the APN. 
The encoder in C-APN was trained using only MSE loss. 
During the inference, the observations given to C-APN as input are obtained by decoding the latent plan found by the LSR with the observation generator $\omega$. \\
\noindent \textbf{R-APN} is an extension of C-APN that 
uses a ResNet encoder identical to the VAE encoder.

Detailed architectures of all the models can be found in our code repository. 
The training details for APN and APN-d are described in Sec.~\ref{sec:folding:details}. For C-APN-d and R-APN-d, we similarly trained $5$ models using different random seeds but on a training dataset $\overline{\mathcal{T}_I}$ obtained by decoding $\overline{\mathcal{T}_z}$ with the observation generator $\omega$ of $\vae{16}{f}{L_1}$. This is because the visual plans, given to C-APN-d and R-APN-d, are produced by decoding the latent plans with $\omega$. Moreover, C-APN-d and R-APN-d were  trained for $1000$ epochs to ensure the convergence of the initialized encoders. Note that we can only obtain one AAB model for a chosen VAE as AAB is defined by the LSR.

We evaluated the performance of all the models on a holdout dataset consisting of $41$ action pairs. Given a holdout action pair, we calculated the mean squared error (MSE) between the predicted and the ground truth action specifics. We report the mean and standard deviation of the obtained MSE calculated across the $5$ random seeds (except for AAB). The results are shown in Table~\ref{tab:apn_basline_comp} where we separately report the error obtained on picking and releasing as well as the total model error.
Firstly, we  
observed that the added regularization layer positively  
 affected the result as APN achieved lower error than our earlier version e-APN~\cite{ouriros}. Secondly, APN significantly outperformed both C-APN and R-APN. Using the latent encodings as inputs also significantly  
 decreased the size of the models and reduces the computational power needed for their training. Lastly,  our APN also on average  outperformed AAB with respect to the total model error.  Although the enhancement compared to the AAB was not as significant as for the other models, APN is beneficial since it is less prone to averaging errors obtained from the LSR and can be easily adapted to any realization of action specifics. Moreover, a neural network realization of the APM potentially allows more accurate modeling of more complex action specifics.  \textit{In summary, using a separate neural network to predict action specifics from latent representations  
led to a lower prediction error and can be easily adapted to different types of actions. }

%-------------------------------------------------
%---------Execution 
%-------------------------------------------------

\subsection{System Analysis}\label{sec:exp:real_robot}
We benchmarked our method against our earlier method in~\cite{ouriros} on the same T-shirt folding task, and additionally measured the performance on a more challenging fold involving picking a layer of the cloth on top of another layer.

%-------------------------------------------------
%---------Folding 
%-------------------------------------------------

\subsubsection{Folding performance and comparison with~\cite{ouriros}}
\label{sec:sec:fold:system_compare}

We performed each fold $5$ times per configuration using the  unseen goal  observations  
shown in Fig.~\ref{fig:folding_s_g} and framework S-OUR, consisting of $\vae{16}{f}{L_1}$, $\lsr{1}$ and $\apn{16}{f}{L_1}$, and
compared the performance with the results from our earlier work~\cite{ouriros} obtained using S-${L_1}$, S-${L_2}$ and S-$L_{\infty}$.

The results are shown in Table \ref{tab:folding:results_folding}, while, as previously mentioned, all execution videos, including the respective visual action plans, are available on the website\footrefc{fn:website}. 
We report the total system success rate with re-planning,  
the percentage of correct single transitions,  
and the  percentage of successful  
visual plans and  action plans from start to goal.  
We observed that S-OUR outperformed the systems from~\cite{ouriros} with a notable $96\%$ system performance, only missing a single folding step which results in a transition performance of $99\%$. As for S-$L_1$, S-OUR also achieved optimal performance when scoring the initial visual plans $P_I$ and the initial action plans $P_u$. 
\textit{We thus conclude that the improved MM, LSR and APM modules 
together contribute to a significant better system than in \cite{ouriros}. }

\begin{table}[]
%\resizebox{\textwidth}{!}{%
\centering
\begin{tabular}{|l|l|l|l|l|}
\hline
Method & Syst. & Trans. & $P_I$ & $Pu$ \\ \hline
\multicolumn{5}{|c|}{Fold 1 to 5 - comparison to \cite{ouriros}} \\ \hline
S-OUR        &   {\boldmath$ 96 \%$}    &       {\boldmath$ 99\%$}       &        {\boldmath$100\%$} &        {\boldmath$100\%$}            \\ \hline
S-$L_1$~\cite{ouriros} &     $80 \%$   &  $90 \%$      &  $100 \%$       &     $100 \%$                 \\ \hline
 S-$L_2$~\cite{ouriros} &      $40 \%$      &  $77 \%$      &    $60 \%$           &       $60 \%$         \\ \hline
S-$L_\infty$~\cite{ouriros} &     $24 \%$    &   $44 \%$    &       $56 \%$       &        $36 \%$         \\ \hline
\multicolumn{5}{|c|}{Fold layer} \\ \hline
S-OUR        &   $ 50\%$    &       $ 83\%$       &        $100\%$  &        $100\%$          \\ \hline
\end{tabular}%
%}
\caption{Results (best in bold) for executing visual action plans on $5$ folding tasks (each repeated $5$ times) shown in the top. The bottom row shows the results on the fold requiring to pick the top layer of the garment (repeated $10$ times). }
\label{tab:folding:results_folding}
\end{table}

%-------------------------------------------------
%---------multiple layers 
%-------------------------------------------------

\subsubsection{Folding with multiple layers}
\label{sec:sec:fold:m_layers}

\begin{figure}[htb!]
\begin{center}
\includegraphics[width=\linewidth]{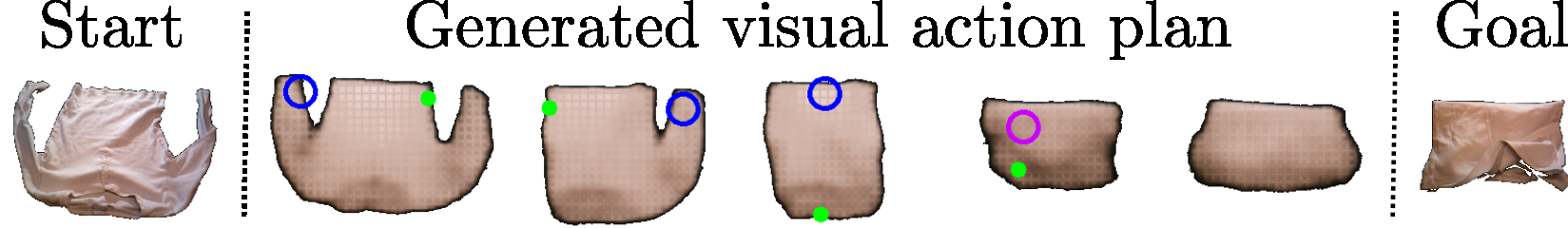}
\end{center}
\caption{Visual action plan for the fold requiring to pick the top layer of the garment. The step where the top layer is to be picked is indicated in purple (see accompanying video for further details).}
\label{fig::layerd_fold}
\end{figure}

As the previous folds resulted in nearly perfect performance of our system, we challenged it with an additional much harder fold that requires to pick the top layer of the garment.  
The fold, shown in Fig.~\ref{fig:folding_s_g} bottom right, was repeated $10$ times. An example of the obtained visual action plan is shown in Fig.~\ref{fig::layerd_fold} and the final results are reported in Table~\ref{tab:folding:results_folding} (bottom row).

Experiments showed that the system had no trouble \textit{planning} the folding steps
 from the initial configuration and   was able to properly plan layer folds (with pick location marked in purple). Concerning the \textit{execution} of the plan,  the robot managed to  correctly fold in $80\%$ of the cases, excluding the last fold, using the re-planning strategy. However, failure cases often occurred  during the execution of the last layer fold, resulting in the robot  picking up multiple layers at the same time. When this happened, the T-shirt deformed into unseen states that were very dissimilar from the ones in $\mathcal{T}_I$ and that rendered the re-planning step inefficient. 
A more precise manipulation system, either using a specialized gripper or custom methods for separating cloth layers, could potentially boost the performance of our system on this specific folding task. We leave these improvements for future work.

\section{Conclusions}\label{sec:concl}

In this work, we presented an extended 
version of the Latent Space Roadmap first introduced in \cite{ouriros} which allows visual action planning of manipulation tasks. Firstly, we improved the building procedure of the LSR in the latent space by 
introducing an outer optimization loop that eliminates the need for a hard-to-tune clustering parameter.
Secondly, we improved the training procedure of the VAE, used to represent the Mapping Module, by dynamically increasing the desired distance between action pairs. We thoroughly investigated the structure of the latent space, and presented a deep insight into the effects that each of the improvements have for the system.  In addition, we compared different realizations of the Action Proposal Module and showcased the benefits of using latent representations for generating action plans.  
Lastly, we evaluated the LSR on three simulated tasks as well as real-world folding task. We introduced a harder version of the box stacking task and a rope-box manipulation task involving a rigid and deformable object, which enabled a more informative ablation study. We showed that the improved LSR significantly outperforms the one presented in~\cite{ouriros} on the same folding task.

We are convinced that in order to advance state-of-the-art manipulation techniques for rigid and deformable objects, improvements on two fronts are necessary: learning a structured latent space as well as its exploration. We believe that our proposed method is a step toward achieving this goal 
which also opens many interesting future directions. For example, we wish to expand our method to encode full trajectories to further structure the latent space, or to apply it to reinforcement learning settings with active exploration.

\begin{appendices}
    \section*{Appendix}

\subsection{Latent space dimension}
\label{sec:sec:stack:ld_dim}

The problem of choosing a suitable latent space dimension has not received much attention in the literature.
% Even though the action term alleviates the problem of indistinguishable distances in higher dimensions by  increasing the relative contrast, it is still important to choose a latent dimension large enough so that the relevant features can be encoded.
In Table \ref{tab:stacking_latent_eval} we report the partial scoring on normal and hard stacking  and rope-box tasks  using VAE models with various latent dimensions. 
The results demonstrate an evident drop in the performance when the latent dimension 
was too small, such as $ld=4$. As $ld$ 
 increased, we  observed gradual improvements in the performance where a satisfactory level  
was achieved using $ld \ge 6$ for $ns$, and $ld \ge 12$ for $hs$  and $rb$. Therefore, $hs$  and $rb$
 required more dimensions in order to capture all the relevant and necessary features. \textit{This result not only demonstrates the complexity of each task version but also justifies the choice $ld = 12$ in the simulations.}
\vspace{11pt}

\begin{table}[h!]
\centering
{ 
\begin{tabular}{|c|c|c|c|c|}
\hline
$ld$ & $ns$ [\%] & $hs$ [\%] & $rb$ [\%]\\ \hline
$4$ & $7.9 \pm 2.2$ & $8.8 \pm 7.9$ &  $62.7 \pm 13.9$\\ \hline
$6$ & $99.96 \pm 0.08$ & $56.2 \pm 23.1$ & $74.9 \pm 5.0$ \\ \hline
$8$ & $99.96 \pm 0.08$ & $62.7 \pm 18.7$ & $80.6 \pm 5.3$  \\ \hline
$12$ & $100.0 \pm 0.0$ & $92.1 \pm 2.9$  & $90.4 \pm 2.9$ \\ \hline
$16$ & $100.0 \pm 0.0$  & $95.9 \pm 1.4$ & $92.2 \pm 1.1$   \\ \hline
$32$ &  $97.5 \pm 4.33$ & $96.4 \pm 0.4$ & $92.6 \pm 2.0$  \\ \hline
\end{tabular}%
%}
}
\caption{Comparison of the LSR performance when using VAEs with different latent dimensions for all the simulated tasks.}
\label{tab:stacking_latent_eval}
\end{table}

\subsection{Overview of clustering algorithms}\label{sec:app:clus-ae} 
In this section, we provide a brief overview of the ablated clustering methods considered in Sec.~\ref{sec:sec:stack:lsr_phase2}. 

\noindent
\textbf{Epsilon clustering:}  used in our earlier work \cite{ouriros} and  functionally coincident with DBSCAN~\cite{ester1996density}. Its performance is affected by the parameter $\epsilon$, \ie, 
radius of the  $\epsilon$-neighborhood of every point, 
and deteriorates when clusters have different densities. 

\noindent
\textbf{Mean-shift:} centroid-based algorithm \cite{cheng1995mean} with moving window approach to identify high density regions. At each iteration, the centroid candidates associated to the windows are updated to the mean of the points in the considered region. The window size has a significant influence on the performance.

\noindent
\textbf{OPTICS:} improved version of DBSCAN introduced by \cite{ankerst1999optics} in which a hierarchical reachability-plot dendrogram is built, whose slope identifies clusters with different densities. 
The parameter $\Xi \in [0,1]$ is used to tune the slope and
 heavily affects the outcome of the algorithm. However,  its influence is not easy to understand intuitively, as discussed in~\cite{campello2013density}.

\noindent
 \textbf{Linkage}:
hierarchical, agglomerative clustering algorithm discussed in Sec.~\ref{ssec:LSR_building}. 
Possible dissimilarity functions to merge points are \textit{single}, based on the minimum distance between any pair of points belonging to two distinct clusters,   \textit{complete}, based on the maximum distance,  and \textit{average}, based on the unweighted average of the distances of all points belonging to two distinct clusters. 

As discussed in Sec.~\ref{ssec:LSR_building}, the clustering threshold $\tau$ determines the vertical cut through the dendrogram and consequently influences the performance of the algorithm.

\noindent
\textbf{HDBSCAN:} 
{agglomerative clustering algorithm in which the branches of the dendrogram are optimized for non-overlapping clusters using a notion of ``cluster stability" based on their longevity. 
HDBSCAN automatically identifies clusters with different densities and requires specifying only the minimum cluster size prior to the training.}

\end{appendices}

\bibliographystyle{ieeetr}
\bibliography{bibliography}

\begin{IEEEbiography}[{\includegraphics[width=1in,clip]{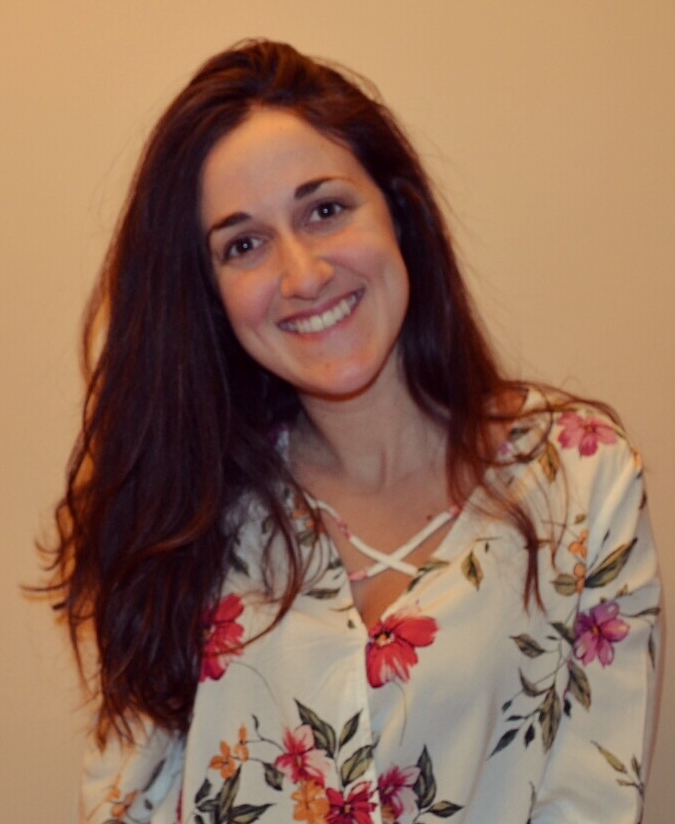}}]{Martina Lippi}
 received the M.Sc. (cum laude) and Ph.D. degrees
 in Information Engineering from the
University of Salerno, Italy, in 
2017 and 2020, respectively.
She has been a Visiting Scholar with the KTH
Royal Institute of Technology,  Sweden,
in 2019. She was a Postdoctoral researcher at Roma Tre University, Italy from November 2020 to June 2022. Since June 2022, she is Assistant Professor at Roma Tre University, Italy.  
Her
research interests include human–robot interaction,
multimanipulator systems, and distributed control.
\end{IEEEbiography}
%\vskip -3\baselineskip

\begin{IEEEbiography}[{\includegraphics[width=1in,height=1.25in,clip]{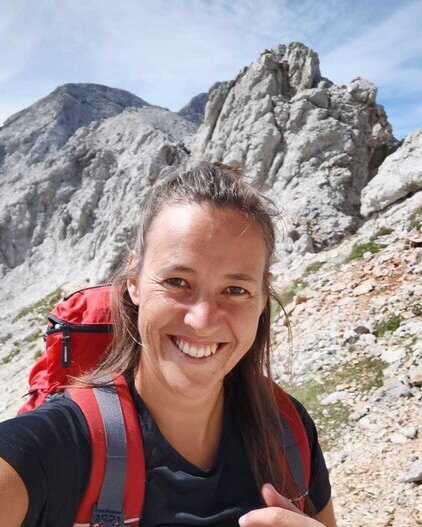}}]{Petra Poklukar}
is a machine learning researcher focusing on representation learning and deep generative models. She received her Master's degree in theoretical mathematics from University of Ljubljana in 2016, and her PhD degree from KTH Royal Institute of Technology in 2022, supervised by Danica Kragic.
\end{IEEEbiography}
%\vskip -3\baselineskip

\begin{IEEEbiography}[{\includegraphics[width=1in,height=1.25in,clip]{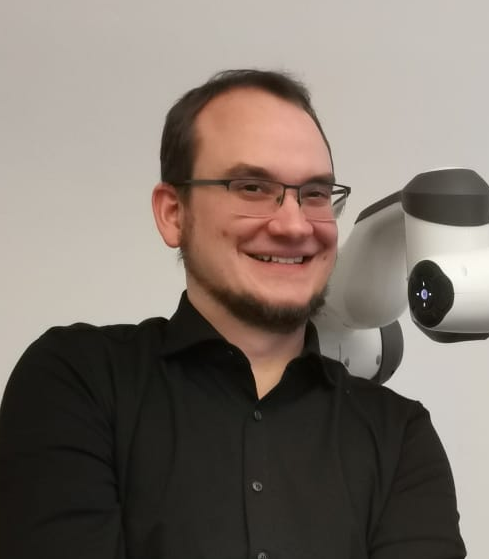}}]{Michael C. Welle}
%Michael C. Welle 
is a Postdoctoral Researcher at KTH Royal Institute of Technology EECS/RPL focusing on representation learning for deformable object manipulation since January 2022.
He obtained his MSc in  Systems, Control and Robotics at KTH in January 2018. 
His subsequent 
Ph.D. research was performed under the supervision of Danica Kragic  
at KTH. 
The title of his thesis is "Learning Structured Representations for Rigid and Deformable Object Manipulation" published in December 2021.
\end{IEEEbiography}
\vskip -3\baselineskip
 
\begin{IEEEbiography}[{\includegraphics[width=1in,height=1.25in,clip]{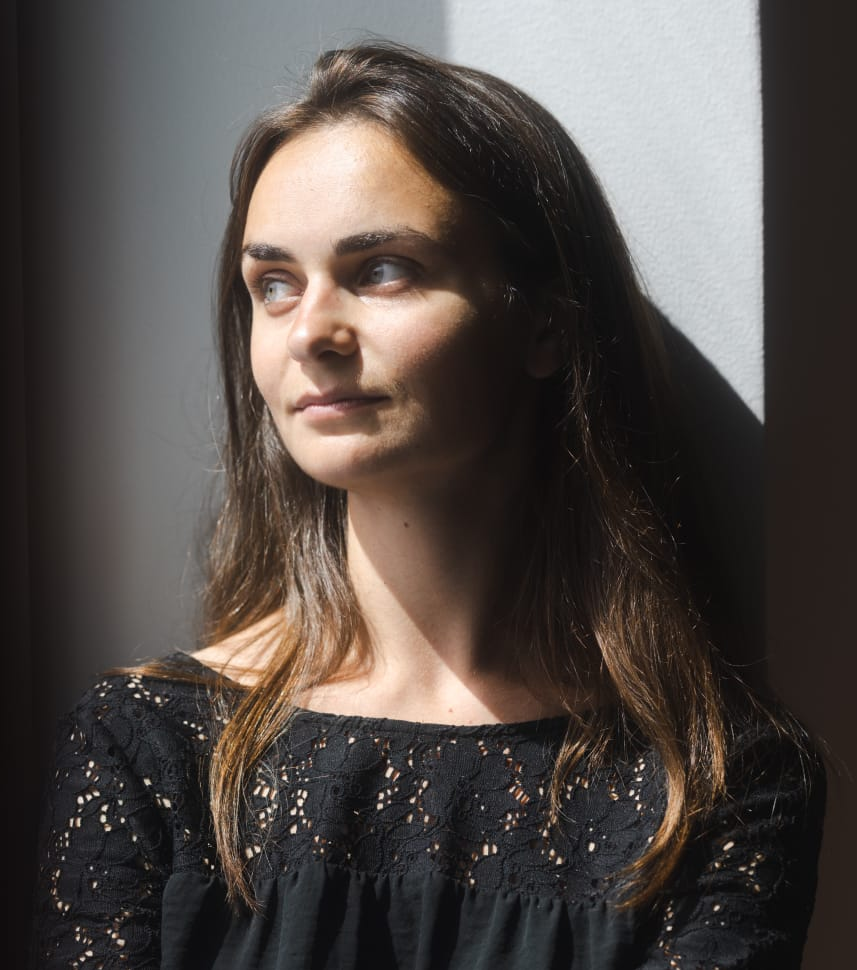}}]{Anastasia Varava}
obtained her PhD in Computer Science from KTH, Sweden in 2019. Her main research interests lie in designing and evaluating efficient representations for various applications, including robotics, molecular science, and social network analysis. She is particularly interested in applying tools and methods from computational geometry and topology to create mathematically rigorous representations and study their properties.
\end{IEEEbiography}
\vskip -3\baselineskip

\begin{IEEEbiography}[{\includegraphics[width=1in,height=1.25in,clip]{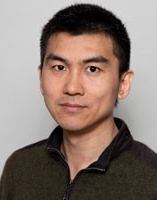}}]{Hang Yin} 
is a postdoctoral researcher with the Division of Robotics, Perception and Learning,  KTH Royal Institute of Technology. He received Bachelor degrees in Mechanical Engineering and Computer Engineering (2007), Master in Mechatronics (2010), both at Shanghai Jiao Tong University, and his PhD degree from Swiss Federal Institute of Technology Lausanne (EPFL) and IST, University of Lisbon (2018). His research interests include modeling, representing, learning and control robot motion and application in human-robot interaction tasks.
\end{IEEEbiography}
\vskip -3\baselineskip

\begin{IEEEbiography}[{\includegraphics[width=1in,height=1.25in,clip]{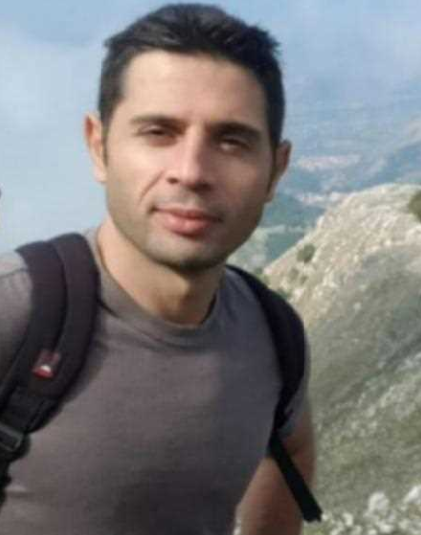}}]{Alessandro Marino}
received the M. Sc. degree cum  laude in Computer Science Engineering from the University of Naples Federico II, Italy, in 2006,  and the Ph.D. degree in automation and robotics from the University of Basilicata, Italy, in 2010.
Since 2018, he is an Associate Professor with the University of Cassino and Southern Lazio. 
His research interests include modeling and control of robotic systems,  multi-robot systems, human-robot-interaction, distributed control.
\end{IEEEbiography}
\vskip -3\baselineskip

\looseness=-1
\begin{IEEEbiography}[{\includegraphics[width=1in,height=1.25in,clip]{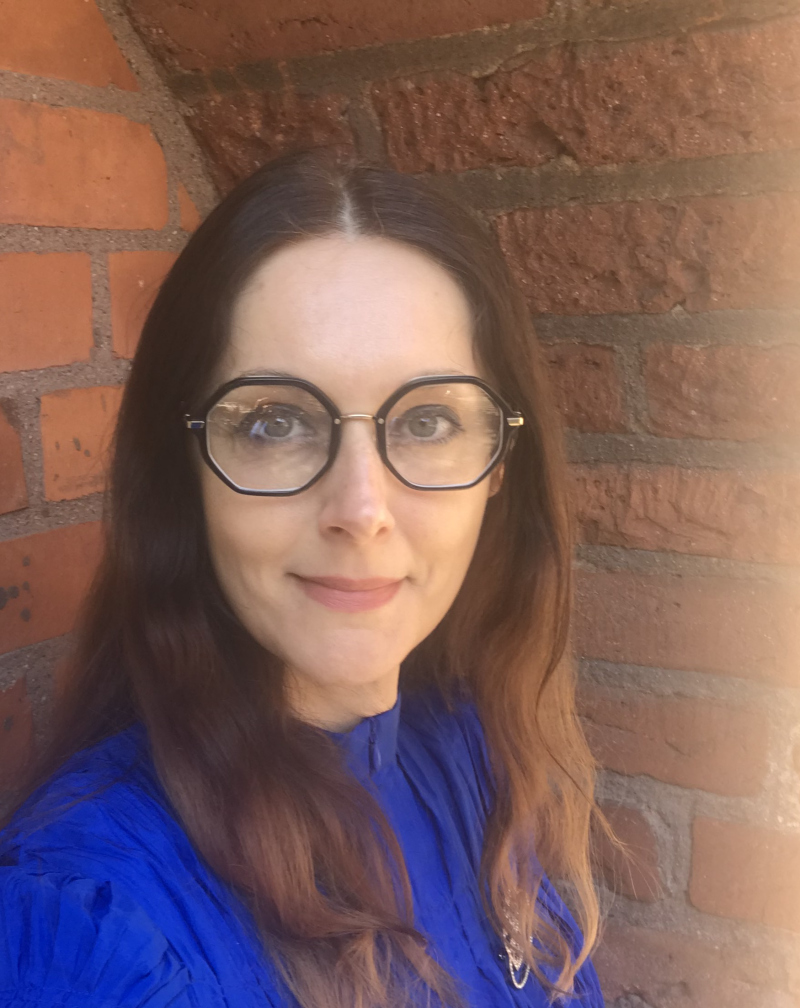}}]{Danica Kragic}
is a Professor at the School of
Electrical Engineering and Computer Science  at KTH in
Stockholm. She received MSc in Mechanical Engineering
from the Technical University of Rijeka,
Croatia in 1995 and PhD in Computer Science
from KTH in 2001. Danica received the 2007 IEEE
Robotics and Automation Society Early Academic
Career Award. She is a member of the Swedish
Royal Academy of Sciences and Swedish Academy of Engineering Sciences.
Her research spans over areas of robotics, machine learning and computer vision.
\end{IEEEbiography}

\end{document}